%% file: main.tex
\RequirePackage{fix-cm}
\documentclass[twocolumn]{svjour3}          
\smartqed  
\usepackage{natbib}
\usepackage{times}
\usepackage{epsfig}
\usepackage{graphicx}
\usepackage{subfigure}
\usepackage{amsmath,amssymb,amsfonts,dsfont,pifont,bm,bbm,mathrsfs,mathtools,nicefrac}
\usepackage{bm}
\usepackage{textcomp}
\usepackage{multirow}
\usepackage{xspace}
\usepackage{xcolor}
\usepackage{booktabs}
\usepackage{tabularx}
\usepackage{color}
\usepackage{makecell}
\usepackage{wrapfig}
\usepackage{bbding}

\usepackage[ruled,linesnumbered]{algorithm2e}

\definecolor{citecolor}{RGB}{119,185,0} 
\usepackage[pagebackref=false,breaklinks=true,colorlinks,citecolor=citecolor,bookmarks=false]{hyperref}

\definecolor{upcolor}{RGB}{57,182,74}

\newlength\savewidth
\definecolor{Light}{RGB}{246,234,227}

\newcommand{\method}{U$^{\text{2}}$PL+\xspace}

\renewcommand{\paragraph}[1]{\vspace{1.25mm}\noindent\textbf{#1}}

\makeatletter
\DeclareRobustCommand\onedot{\futurelet\@let@token\@onedot}
\def\@onedot{\ifx\@let@token.\else.\null\fi\xspace}

\usepackage{soul}
\soulregister\cite7
\soulregister\eqref7
\soulregister\ref7

\begin{document}

\title{Using Unreliable Pseudo-Labels for Label-Efficient Semantic Segmentation
}


\author{Haochen~Wang$^{1,2}$ \and
        Yuchao~Wang$^4$ \and
        Yujun~Shen$^5$ \and
        Junsong~Fan$^3$ \and
        Yuxi~Wang$^3$ \and 
        Zhaoxiang~Zhang$^{1,2,3*}$
}


\institute{
    $^1$ New Laboratory of Pattern Recognition, State Key Laboratory of Multimodal Artificial Intelligence Systems, Institute of Automation, Chinese Academy of Sciences, Beijing, China.
    \\
    $^2$ University of Chinese Academy of Sciences, Beijing, China. \\
    $^3$ Centre for Artificial Intelligence and Robotics, Hong Kong Institute of Science \& Innovation, Chinese Academy of Sciences, Hong Kong, China. \\  
    $^4$ Shanghai Jiao Tong University, Shanghai, China. \\
    $^5$ Chinese University of Hong Kong, Hong Kong, China. \\
    $^*$ Zhaoxiang Zhang is the corresponding author. \\
    E-mail: \{wanghaochen2022, zhaoxiang.zhang\}@ia.ac.cn.
}

\date{Received: date / Accepted: date}

\maketitle

\begin{abstract}
The crux of label-efficient semantic segmentation is to produce high-quality pseudo-labels to leverage a large amount of unlabeled or weakly labeled data.
%
%
A common practice is to select the highly confident predictions as the pseudo-ground-truths for each pixel, but it leads to a problem that most pixels may be left unused due to their unreliability.
However, we argue that \textit{every pixel matters to the model training}, even those unreliable and ambiguous pixels.
Intuitively, an unreliable prediction may get confused among the top classes, however, it should be confident about the pixel not belonging to the remaining classes.
Hence, such a pixel can be convincingly treated as a negative key to those most unlikely categories.
Therefore, we develop an effective pipeline to make sufficient use of unlabeled data.
Concretely, we separate reliable and unreliable pixels via the entropy of predictions, push each unreliable pixel to a category-wise queue that consists of negative keys, and manage to train the model with all candidate pixels.
Considering the training evolution, we adaptively adjust the threshold for the reliable-unreliable partition.
Experimental results on various benchmarks and training settings demonstrate the superiority of our approach over the state-of-the-art alternatives.
\end{abstract}

\input{1.intro}

\input{2.related}
\input{3.method}

\input{4.exp}

\input{5.conclusion}


\section*{Declarations}

\noindent\textbf{Acknowledgements}
This work was supported in part by the National Key R\&D Program of China (No. 2022ZD0116500), the National Natural Science Foundation of China (No. U21B2042), and in part by the 2035 Innovation Program of CAS, and the InnoHK program.

\noindent\textbf{Data Availability.}
The datasets generated during and/or analyzed during the current study are available from the PASCAL VOC 2012\footnote{http://host.robots.ox.ac.uk/pascal/VOC/}, the SBD\footnote{https://ieeexplore.ieee.org/abstract/document/6126343}, the GTA5\footnote{https://arxiv.org/pdf/1608.02192v1.pdf}, the Synthia\footnote{https://synthia-dataset.net/}, and the Cityscapes\footnote{https://www.cityscapes-dataset.com/}.

%
%


{\footnotesize
    \bibliographystyle{apalike}
    \bibliography{ref.bib}
}

\end{document}

%% file: 1.intro.tex
\section{Introduction}\label{sec:introduction}

\begin{figure}[t]
    \centering
    \includegraphics[width=1.0\linewidth]{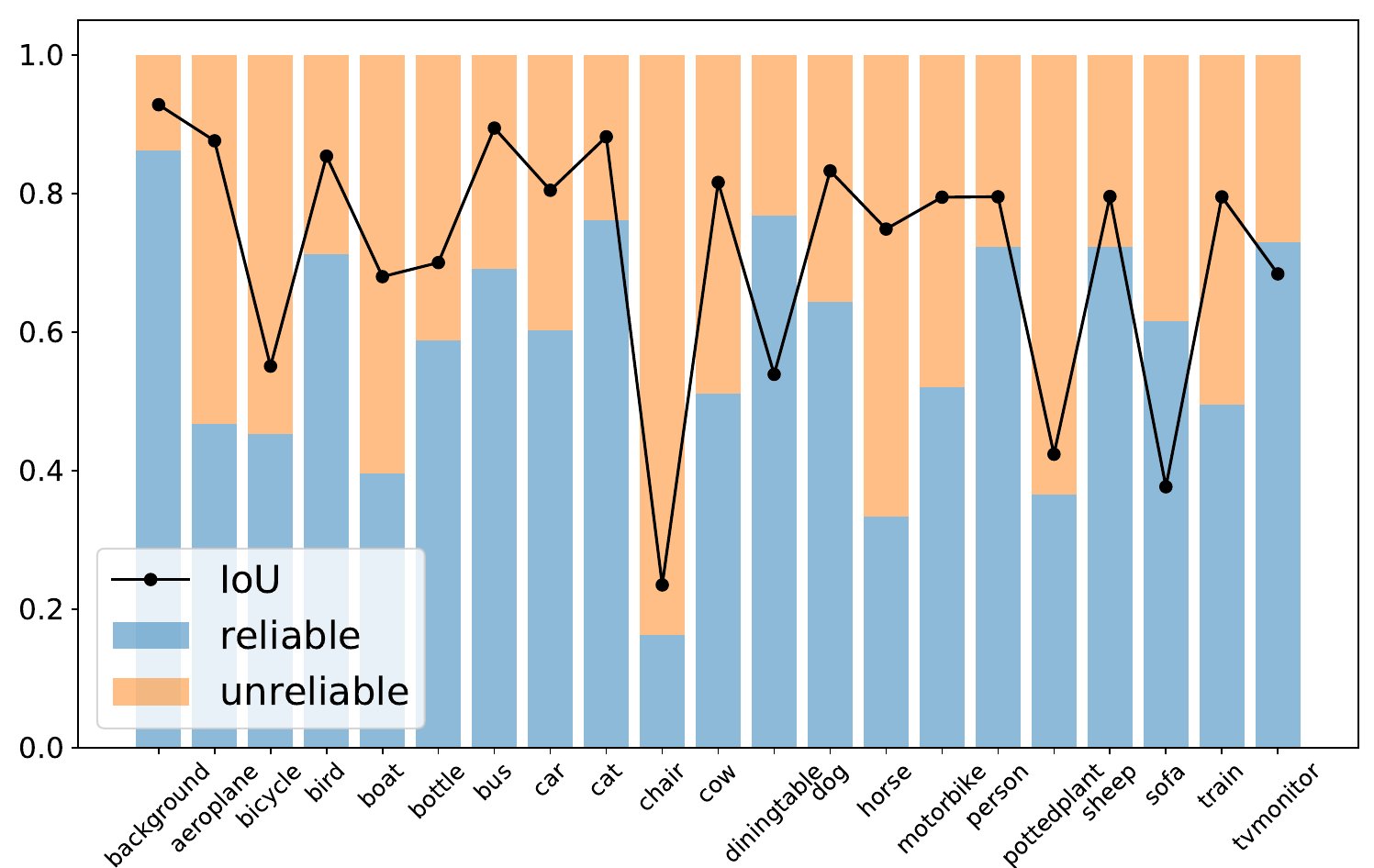}
    \vspace{-10pt}
    \caption{
        \textbf{Category-wise performance and statistics on the number of pixels with reliable and unreliable predictions.}
        Model is trained using $732$ labeled images on PASCAL VOC 2012 \citep{voc} and evaluated on the remaining $9,850$ images.
    }
    \label{fig:stats}
    \vspace{-10pt}
\end{figure}

Semantic segmentation is a fundamental task in the computer vision field and has been significantly advanced along with the rise of deep neural networks \citep{fcn, unet, pspnet, deeplab}.
However, existing supervised approaches rely on large-scale annotated data, which can be too costly to acquire in practice.
For instance, it takes around 90 minutes to annotate just a single image of Cityscapes \citep{cityscapes}, and this number even increases to 200 under adverse conditions \citep{sakaridis2021acdc}.
%
%
To alleviate this problem, many attempts have been made towards label-efficient semantic segmentation \citep{ru2022learning, wang2022semi, hoyer2022daformer}.
Under this setting, self-training, \textit{i.e.}, assigning pixel-level \textit{pseudo}-labels for each weakly-labeled sample, becomes a typical solution.
Specifically, given a weakly-labeled image, prior arts \citep{lee2013pseudo, xie2020self} borrow predictions from the model trained on labeled data or leverage Class Activation Maps (CAMs) \citep{zhou2016learning} to obtain pixel-wise prediction, and then use them as the ``ground-truth'' to, in turn, boost the model.
Along with it, many attempts have been made to produce high-quality pseudo-labels.
A typical solution is to filter the predictions using their confidence scores \citep{st++, pseudoseg, zuo2021self, dash, hoyer2022daformer, ru2022learning}.
In this way, \textit{only} the highly confident predictions are served as the pseudo-labels, while \textit{those ambiguous ones are simply discarded.}

One potential problem caused by this paradigm is that some pixels may \textit{never} be learned in the entire training process.
For example, if the model cannot satisfyingly predict some certain class, it becomes difficult to assign accurate pseudo-labels to the pixels regarding such a class, which may lead to insufficient and categorically imbalanced training.
For instance, as illustrated in Fig.~\ref{fig:stats}, underperformed category \texttt{chair} tends to have fewer reliable predictions, and the model will be probably biased to dominant classes, \textit{e.g.}, \texttt{background} and \texttt{cat}, when we filter reliable predictions to be ground-truths for supervision and simply discard those ambiguous predictions.
This issue even becomes more severe under domain adaptive or weakly supervised settings.
Under both settings, predictions of unlabeled or weakly-labeled images usually run into undesirable chaos, and thus only a few pixels can be regarded as reliable ones when selecting highly confident pseudo-labels.
%
%
From this perspective, we argue that to make full use of the unlabeled data, every pixel should be properly utilized.

However, how to use these unreliable pixels appropriately is non-trivial.
Directly using the unreliable predictions as the pseudo-labels will cause the performance degradation \citep{arazo2020pseudo} because it is almost impossible for unreliable predictions to assign exactly correct pseudo-labels.
Therefore, in this paper, we propose an alternative way of Using Unreliable Pseudo-Labels (U$^2$PL).

\begin{figure}[t]
    \centering
    \includegraphics[width=1.0\linewidth]{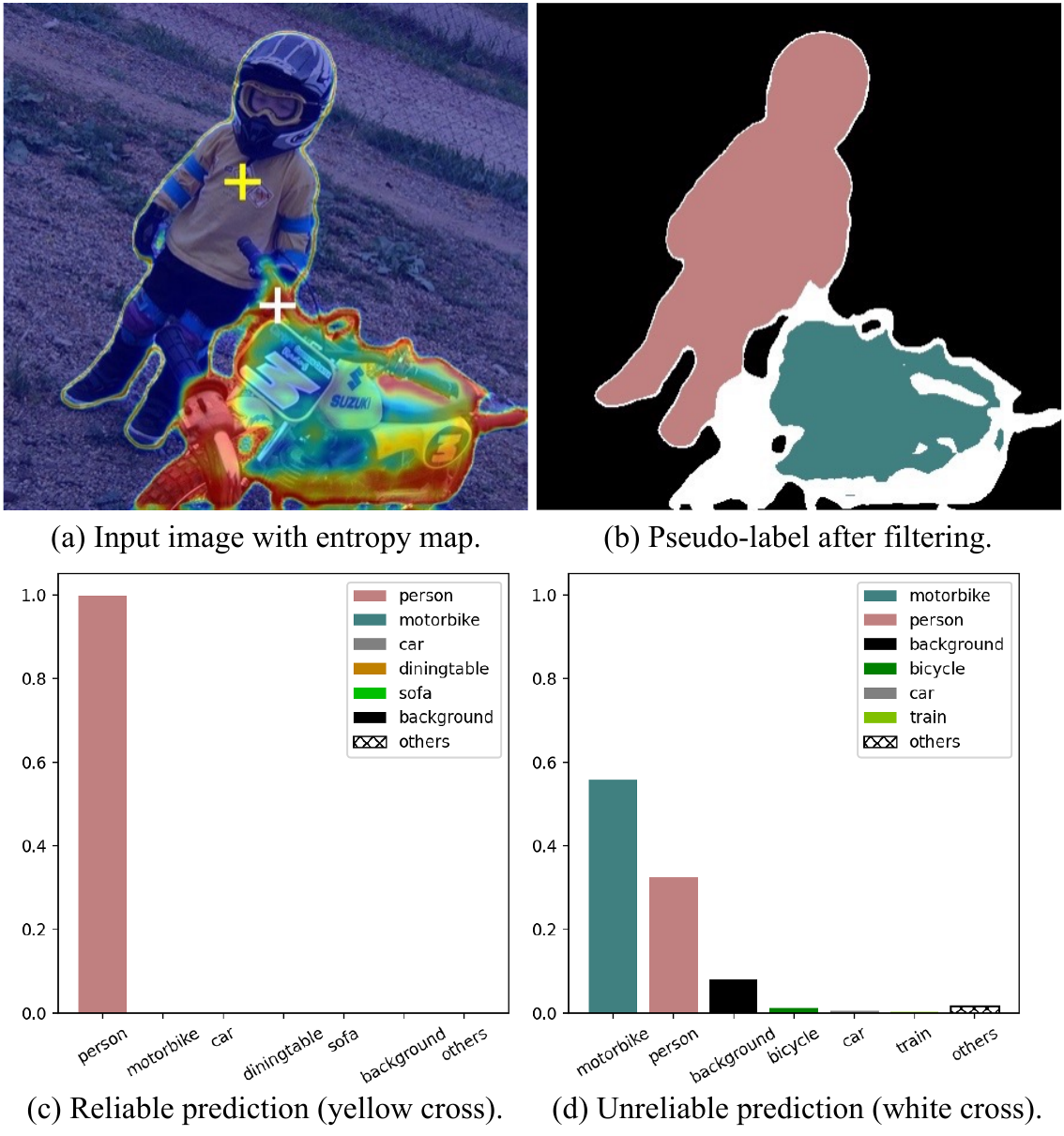}
    \vspace{-15pt}
    \caption{
        \textbf{Illustration on unreliable pseudo-labels.}
        (a) Pixel-wise entropy predicted from an unlabeled image.
        %
        (b) Pixel-wise pseudo-labels from reliable predictions \textit{only}, where pixels within the white region are not assigned a pseudo-label.
        (c) Category-wise probability of a reliable prediction (\textit{i.e.}, the yellow cross).
        %
        (d) Category-wise probability of an unreliable prediction (\textit{i.e.}, the white cross), which hovers between \texttt{motorbike} and \texttt{person}, yet is confident enough of \textit{not} belonging to \texttt{car} and \texttt{train}.
    }
    \label{fig:example}
    \vspace{-10pt}
\end{figure}

First, we observe that an unreliable prediction usually gets confused among \textit{only a few} classes instead of all classes.
Taking Fig.~\ref{fig:example} as an instance, the pixel with a white cross is an unreliable prediction that receives similar probabilities on class \texttt{motorbike} and \texttt{person}, but the model is pretty sure about this pixel \textit{not} belonging to class \texttt{car} and \texttt{train}.
Based on this observation, we reconsider those unreliable pixels as the negative keys to those unlikely categories, which is a simple and intuitive way to make full use of all predictions.
Specifically, after getting the prediction from an unlabeled image, we first leverage the per-pixel entropy as the metric (see Fig.~\ref{fig:example}\textcolor{red}{a}) to separate all pixels into two groups, \textit{i.e.}, reliable ones and unreliable ones.
All reliable predictions are then used to derive positive pseudo-labels, while the pixels with unreliable predictions are pushed into a memory bank, which is full of negative keys.
To avoid all negative pseudo-labels only coming from a subset of categories, we employ a queue for each category.
Such a design ensures that the number of negative keys for each class is balanced, preventing being overwhelmed by those dominant categories.
Meanwhile, considering that the quality of pseudo-labels becomes higher as the model gets more and more accurate, we come up with a strategy to adaptively adjust the threshold for the partition of reliable and unreliable pixels.

\paragraph{Extensions of the conference version \citep{wang2022semi}.}
To better demonstrate the efficacy of using unreliable pseudo-labels, instead of studying only under the semi-supervised setting, we extend our original conference publication \citep{wang2022semi} to domain adaptive and weakly-supervised settings, indicating that using unreliable pseudo-labels is crucial and effective on various label-efficient settings, bringing significant improvements \textit{consistently}.
Moreover, to produce high-quality pseudo-labels, 
we further make the category-wise prototype momentum updated during training to build a consistent set of keys and propose a denoising technique to enhance the quality of pseudo-labels.
Additionally, a symmetric cross-entropy loss \citep{wang2019symmetric} is used considering the pseudo-labels are still noisy even after filtering and denoising.
We call the extended framework as \method.

In the following, we provide a brief discussion of how these three settings differ.
Semi-supervised (SS) approaches aim to train a segmentation model with only a few labeled pixel-level ground-truths \citep{st++, cps, chen2021semisupervised, alonso2021semi, french2019semi, cct, wang2022semi} together with numerous unlabeled ones.
Domain adaptive (DA) alternatives \citep{hoyer2022daformer, zhang2021prototypical, li2022class, hoffman2018cycada, wang2020classes, wang2023pulling} introduce synthetic (source) datasets \citep{richter2016gta, ros2016synthia} into training and try to generalize the segmentation model to real-world (target) domains \citep{cityscapes, sakaridis2021acdc} without access to target labels.
Because in most cases, dense labels for those synthetic datasets can be obtained with minor effort.
Weakly-supervised (WS) methods \citep{ru2022learning, fan2020cian, fan2020employing, fan2020learning, li2022towards, zhang2020reliability} leverage weak supervision signals that are easier to obtain, such as image labels \citep{papandreou2015weakly}, bounding boxes \citep{dai2015boxsup}, points \citep{fan2022pointly}, scribbles \citep{lin2016scribblesup}, instead of pixel-level dense annotations to train a segmentation model.

We evaluate the proposed \method on both (1) SS, (2) DA, and (3) WS semantic segmentation settings, where \method manages to bring significant improvements consistently over baselines.
In SS, we evaluate our \method on PASCAL VOC 2012 \citep{voc} and Cityscapes \citep{cityscapes} under a wide range of training settings.
In DA, we evaluate our \method on two widely adopted benchmarks, \textit{i.e.}, GTA5 \citep{richter2016gta} $\to$ Cityscapes \citep{cityscapes} and SYNTHIA \citep{ros2016synthia} $\to$ Cityscapes \citep{cityscapes}.
In WS, we evaluate our \method on PASCAL VOC 2012 \citep{voc} benchmark using only image-level supervisions.
Furthermore, through visualizing the segmentation results, we find that our method achieves much better performance on those ambiguous regions (\textit{e.g.}, the border between different objects), thanks to our adequate use of the unreliable pseudo-labels.
Our contributions are summarized as follows:
\begin{enumerate}
\item
Based on the observation that unreliable predictions usually get confused among only a few classes instead of all classes, we build an intuitive framework U$^2$PL that aims to mine the inherited information of discarded unreliable keys.
\item
We extend the original version of U$^2$PL to  (1) domain adaptive and (2) weakly supervised semantic segmentation settings, demonstrating that using unreliable pseudo-labels is crucial in \textit{both} settings.
\item 
To produce high-quality pseudo-labels, we further incorporate three carefully designed techniques, \textit{i.e.}, momentum prototype updating, prototypical denoising, and symmetric cross-entropy loss.
\item
\method outperforms previous methods across extensive settings on both SS, DA, and WS benchmarks.
\end{enumerate}

%% file: 2.related.tex
\section{Related Work}\label{sec:related}
\noindent\textbf{Semantic segmentation} 
aims to assign each pixel a pre-defined class label, and tremendous success in segmentation brought by deep convolutional neural networks (CNNs) has been witnessed \citep{unet, fcn, yu2015multi, deeplab, chen2017rethinking, pspnet, deeplabv3p, badrinarayanan2017segnet}.
Recently, Vision Transformers (ViTs) \citep{dosovitskiy2021image} provides a new feature extractor for images, and researchers have successfully demonstrated the feasibility of using ViTs in semantic segmentation \citep{zheng2021rethinking, xie2021segformer, cheng2021per, strudel2021segmenter, cheng2022masked, xu2022groupvit}.
However, despite the success of these deep models, they usually thrive with dense per-pixel annotations, which are extremely expensive and laborious to obtain \citep{cityscapes, sakaridis2021acdc}.

\paragraph{Semi-supervised semantic segmentation}
methods aim to train a segmentation model with only a few labeled images and a large number of unlabeled images.
There are two typical paradigms for semi-supervised learning: consistency regularization \citep{bachman2014learning, cct, french2019semi, sajjadi2016regularization, dash} and entropy minimization \citep{em, chen2021semisupervised}. 
Recently, a variant framework of entropy minimization, \textit{i.e.}, self-training \citep{lee2013pseudo}, has become the mainstream thanks to its simplicity and efficacy.
On the basis of self-training, several methods \citep{french2019semi, yuan2021simple, st++, wang2023balancing, du2022learning} further leverage strong data augmentation techniques such as \citep{cutout, cutmix, classmix}, to produce meaningful supervision signals.
However, in the typical weak-to-strong self-training paradigm \citep{fixmatch}, unreliable pixels are usually simply discarded.
\method, on the contrary, fully utilizes those discarded unreliable pseudo-labels, contributing to boosted segmentation results.

\paragraph{Domain adaptive semantic segmentation} 
focuses on training a model on a labeled source (synthetic) domain and generalizing it to an unlabeled target (real-world) domain.
This is a more complicated task compared with semi-supervised semantic segmentation due to the domain shift between source and target domains.
To overcome the domain gap, most previous methods optimize some custom distance \citep{long2015learning, lee2019sliced, wang2023pulling} or apply adversarial training \citep{goodfellow2014generative, nowozin2016f},
in order to align distributions at the image level \citep{hoffman2018cycada, murez2018image, sankaranarayanan2018learning, li2019bidirectional, gong2019dlow, choi2019self, wu2019ace, abramov2020keep, zhang2020cross}, 
intermediate feature level \citep{hoffman2016fcns, hong2018conditional, hoffman2018cycada, saito2018maximum, chang2019all, chen2019progressive, wan2020bringing, li2021t, wang2023pulling}, 
or output level \citep{tsai2018learning, luo2019taking, melas2021pixmatch}.
Few studies pay attention to unreliable pseudo-labels under this setting.
To the best of our knowledge, we are the first to recycle those unreliable predictions when there exists a distribution shift between different domains.

\paragraph{Weakly supervised semantic segmentation} seeks to train semantic segmentation models using only \textit{weak} annotations, and can be mainly categorized into image-level labels \citep{du2022weakly, ru2022learning, ru2023token, fan2020cian, fan2020learning, ahn2018learning, lee2021railroad, wu2021embedded, li2021pseudo, lee2019ficklenet, lee2021anti}, points \citep{fan2022pointly}, scribbles \citep{lin2016scribblesup}, and bounding boxes \citep{dai2015boxsup}.
This paper mainly discusses the image-level supervision setting, which is the most challenging among all weakly supervised scenarios.
Most methods \citep{wei2017object, zhang2021complementary, sun2021ecs, jiang2019integral, kim2021discriminative, yao2021non} are designed with a \textit{multi-stage} process, where a classification network is trained to produce the initial pseudo-masks at pixel level using CAMs \citep{zhou2016learning}.
This paper focuses on \textit{end-to-end} frameworks \citep{pinheiro2015image, papandreou2015weakly, roy2017combining, zhang2020reliability, araslanov2020single} in weakly supervised semantic segmentation with the goal of making full use of pixel-level predictions, \textit{i.e.}, CAMs.



\paragraph{Contrastive learning} is widely used by many successful works in unsupervised visual representation learning \citep{simclrv2, mocov3, wang2023hard, wang2023droppos}.
In semantic segmentation, contrastive learning has become a promising new paradigm \citep{reco, wang2021exploring, zhao2021contrastive}. 
The following methods try to go deeper by adopting the contrastive learning framework for semi-supervised semantic segmentation tasks.
\citep{pc2seg} minimizes the mean square error between two positive samples and introduces several strategies to sample negative pixels. 
\citep{alonso2021semi} utilizes a class-wise memory bank to store representative negative pixels for each class. 
However, these methods ignore the common \textit{false negative samples} in semi-supervised segmentation, where unreliable pixels may be wrongly pushed away in a contrastive loss.
Based on the observation that unreliable predictions usually get confused among only a few categories, \method alleviates this problem by discriminating the unlikely categories of unreliable pixels.
In the field of domain adaptive semantic segmentation, only a few methods apply contrastive learning.
\citep{kang2020pixel} adopts pixel-level cycle association in conducting positive pairs.
\citep{zhou2021domain} apply regional contrastive consistency regularization.
\citep{wang2023pulling} introduces an image translation engine to ensure cross-domain positive pairs are matched precisely.
The underlying motivation of these methods is to build a category-discriminative target representation space.
However, we focus on how to make full use of unreliable pixels, which is quite different from existing contrastive learning-based DA alternatives.
Contrastive learning is also studied in weakly supervised semantic segmentation.
For instance, \citep{du2022weakly} proposes pixel-to-prototype contrast to improve the quality of CAMs by pulling pixels
close to their positive prototypes.
\citep{ru2023token} extend this idea by incorporating the self-attention map of ViTs.
On the contrary, the goal of using contrastive learning in \method is \textit{not} to improve the quality of CAMs.
The contrast of \method is conducted after CAM values are obtained with the goal of fully using unreliable predictions.

{\color{black}

\paragraph{Segment Anything Model} (SAM)~\citep{kirillov2023segment} shows strong generalization capabilities by training on over 1 billion object masks.
However, it is important to note that SAM is not designed for \textit{semantic} segmentation.
Instead, it produces \textit{binary} masks within an image.
Given the impracticality of manually classifying each object mask from the SA-1B~\citep{kirillov2023segment} dataset into specific categories, label-efficient semantic segmentation is still worth studying.
Moreover, only $\approx$1\% masks of SA-1B are manually annotated.
The authors have leveraged a segmentation model to assist in data collection and enhance mask diversity.
They even include a fully automated annotation phase, where the model generates masks without any human input, accounting for 99.1\% of the masks.
Based on this, we hypothesize that an appropriate self-training pipeline might contribute to a more powerful segmentation model.
It is also worth noting that SAM is not universally effective across all domains.
For instance, it fails in medical segmentation and camouflaged object segmentation~\citep{ji2023segment}.
To adapt SAMs to specific domains, additional domain-specific annotations are typically required.
Specifically, \cite{ma2024segment} combined a large amount of public-available medical segmentation datasets and built a dataset with over 1.5M image-mask pairs, which is much less diverse compared with the SA-1B dataset \textit{due to the absence of unlabeled images}.
We believe that developing an efficient pipeline for adapting SAM to specific domains, leveraging unlabeled data with \textit{minimal} annotation costs, is also a worthwhile direction for future research.
}


%% file: 3.method.tex
\section{Method}\label{sec:method}

\begin{figure*}[t]
    \centering
    \includegraphics[width=0.9\textwidth]{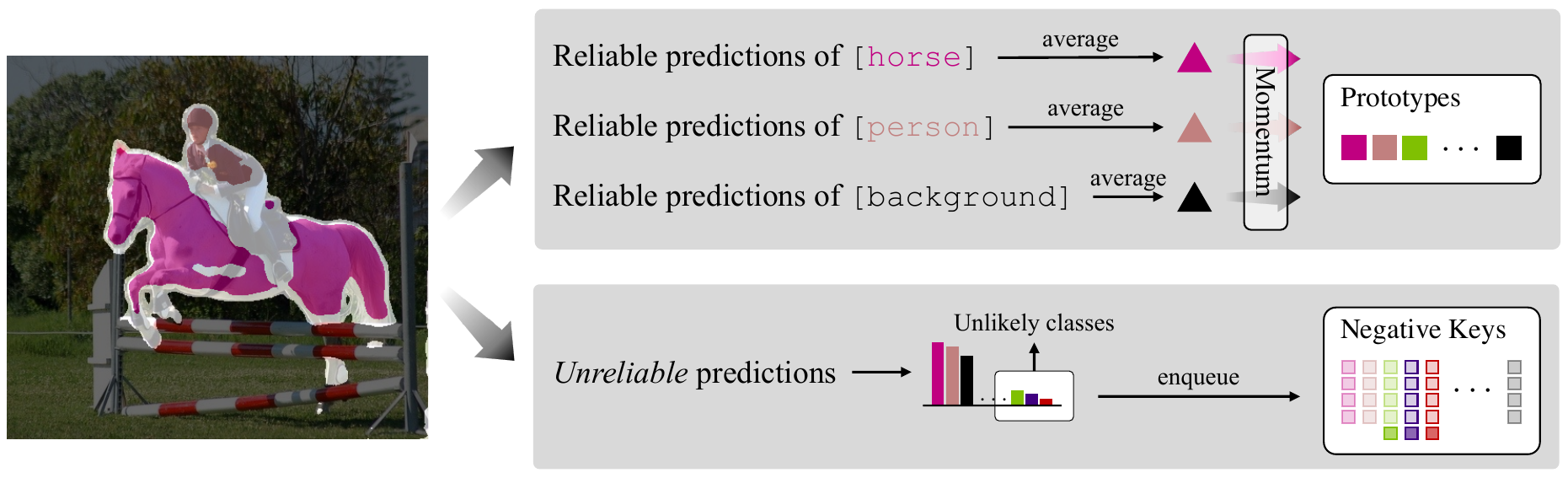}
    \caption{
        \textbf{Illustration of \method.} 
        Segmentation predictions are first split into reliable ones and unreliable ones based on their pixel-level entropy.
        The reliable predictions are used to be the pseudo-labels and to compute category-wise prototypes.
        Each unreliable prediction is pushed into a category-wise memory bank and regarded as negative keys for its unlikely classes.
        Pixels in each memory bank are regarded as the negative samples to the corresponding class, which is formulated as Eq.~(\ref{eq:contraloss}).
    }
    \label{fig:pipeline}
    \vspace{-10pt}
\end{figure*}

In this section, we first introduce background knowledge of label-efficient semantic segmentation in Sec.~\ref{sec:preliminary}.
Next, the elaboration of \method is described in Sec.~\ref{sec:elaboration}.
In Sec.~\ref{sec:pseudo} and Sec.~\ref{sec:denoising}, we introduce how to filter high-quality pseudo-labels and the denoising technique, respectively.
Then, in Sec.~\ref{sec:contra}, we specify how \method can be used in three label-efficient learning tasks, \textit{i.e.}, semi-supervised/domain adaptive, and weakly supervised settings.

\subsection{Preliminaries}
\label{sec:preliminary}

\noindent\textbf{Label-efficient semantic segmentation.}
The common goal of label-efficient semantic segmentation methods is leveraging the \textit{incomplete} set of labels $\mathcal{Y} = \{\mathbf{y}_i\}_{i=1}^{N_l}$ to train a model that segments well on the whole set of images $\mathcal{X} = \{\mathbf{x}_i\}_{i=1}^N$, where $N$ indicates the total images and $N_l$ is the number of labeled samples.
$\mathbf{x}_i \in \mathbb{R}^{H \times W \times 3}$ where $H$ and $W$ are input resolutions.
The first $N_l$ image samples $\mathcal{X}_l = \{\mathbf{x}^l_i\}_{i=1}^{N_l}$ are matched with unique labels, and they consist of the labeled set $\mathcal{D}_l=\left\{(\mathbf{x}_i^l, \mathbf{y}_i^l)\right\}_{i=1}^{N_l}$, while the remaining samples consist of the unlabeled set $\mathcal{D}_u=\left\{\mathbf{x}_i^u\right\}_{i=1}^{N_u}$.

In practice, usually, we have $N_l \ll N$ and each label $\mathbf{y}_i \in \{0,1\}^{H \times W \times C}$ is the one-hot pixel-level annotation, where $C$ is the number of categories.
The labeled set is collected from distribution $\mathcal{P}$ while the unlabeled set is sampled from distribution $\mathcal{Q}$.
A general case is $\mathcal{P} \neq \mathcal{Q}$, and thus the problem falls into \textit{domain adaptive} semantic segmentation. 
Otherwise, if $\mathcal{P} = \mathcal{Q}$, it is usually treated as a \textit{semi-supervised} semantic segmentation task.

However, when $\mathcal{P} = \mathcal{Q}$ but those labels are not collected at pixel-level, it becomes \textit{weakly supervised} semantic segmentation.
Under such a setting, each image $\mathbf{x}_i$ is matched with its corresponding weak label $\mathbf{y}_i$, such as image labels \citep{papandreou2015weakly}, bounding boxes \citep{dai2015boxsup}, points \citep{fan2022pointly}, and scribbles \citep{lin2016scribblesup}, and thus $N=N_l$.
This paper studies the image-level supervision setting, \textit{i.e.}, $\mathbf{y}_i \in \{0,1\}^{C}$, which is the most challenging scenario.
Note that $\mathbf{y}_i$ is not the one-hot label since each image usually contains more than one category.

The overall objective of both settings usually contains a supervised term $\mathcal{L}_s$ and an unsupervised term $\mathcal{L}_u$:
\begin{equation}
\label{eq:common}
    \mathcal{L} = \mathcal{L}_s + \lambda_u \mathcal{L}_u,
\end{equation}
where $\lambda_u$ is the weight of the unsupervised loss, which controls the balance of these two terms.
Next, we will introduce the conventional pipeline of different settings, respectively.

\paragraph{Semi-supervised and domain adaptive semantic segmentation.}
The typical paradigm of these two settings is the self-training framework \citep{meanteacher, fixmatch}, which consists of two models with the same architecture, named \textit{teacher} and \textit{student}, respectively.
These two models differ only when updating their weights.
$\theta_s$ indicates the weight of the student and is updated consistently with the common practice using back-propagation, while the teacher's weights $\theta_t$ are exponential moving average (EMA) updated by the student's weights: 
\begin{equation}
    \theta_t \leftarrow m\theta_t + (1-m)\theta_s,
\end{equation}
where $m$ is the momentum coefficient.

Each model consists of an encoder $h$, a segmentation head $f$, and a representation head $g$ used only for \method.
At each training step, we equally sample $B$ labeled images $\mathcal{B}_l$and $B$ unlabeled images $\mathcal{B}_u$, and try to minimize Eq.~(\ref{eq:common}).
Mathematically, $\mathcal{L}_s$ is the vanilla pixel-level cross-entropy:
\begin{equation}
\label{eq:suploss}
\begin{aligned}
    &\mathcal{L}_s = \frac{1}{|\mathcal{B}_l|} \sum_{(\mathbf{x}_i^l, \mathbf{y}_i^l) \in \mathcal{B}_l} \ell_{ce}(f\circ h(\mathbf{x}_i^l; \theta), \mathbf{y}_i^l), \\
    &\ell_{ce}(\mathbf{p}, \mathbf{y}) = - \mathbf{y}^\top \log \mathbf{p},
\end{aligned}
\end{equation}
where $\mathbf{x}_i^l$ indicates the $i$-th labeled image and $\mathbf{y}_i^l$ represents the corresponding one-hot hand-annotated segmentation map.
$f\circ h$ is the composition function of $h$ and $f$, which means the images are first fed into $h$ and then $f$ to get segmentation results.
As the unsupervised term $\mathcal{L}_u$, we set it as the symmetric cross-entropy (SCE) loss \citep{wang2019symmetric} for a stable training procedure, especially in the early stage.
When computing $\mathcal{L}_u$, we first take each unlabeled sample $\mathbf{x}^u_i$ into the teacher model and get predictions.
Then, based on the pixel-level entropy map, we ignore unreliable pseudo-labels.
Specifically,
\begin{equation}
\begin{aligned}
\label{eq:unsloss}
    &\mathcal{L}_u = \frac{1}{|\mathcal{B}_u|} \sum_{\mathbf{x}_i^u \in \mathcal{B}_u} \ell_{sce}(f\circ h(\mathbf{x}_i^u; \theta), \tilde{\mathbf{y}}_i^u), \\
    &\ell_{sce}(\mathbf{p}, \mathbf{y}) = \xi_{1}\ell_{ce}(\mathbf{p}, \mathbf{y}) + \xi_{2}\ell_{ce}(\mathbf{y}, \mathbf{p}),
\end{aligned}
\end{equation}
where $\tilde{\mathbf{y}}_i^u$ is the one-hot pseudo-label for the $i$-th unlabeled image.
We set $\xi_1$ to $1$ and $\xi_2$ to $0.5$ follow \citep{wang2019symmetric} and \citep{zhang2021prototypical}.

By minimizing $\mathcal{L}_s$ and $\mathcal{L}_u$ \textit{simultaneously}, the model is able to leverage both the small labeled set $\mathcal{D}_l$ and the larger unlabeled set $\mathcal{D}_u$.

\paragraph{Weakly supervised semantic segmentation using image-level labels.}
Weakly supervised semantic segmentation methods aim to leverage CAMs \citep{zhou2016learning} to produce pixel-level pseudo-masks first and then train a segmentation module synchronously \citep{ru2022learning, araslanov2020single}, \textit{i.e.}, end-to-end, or asynchronously \citep{du2022weakly, lee2021railroad}, \textit{i.e.}, multi-stage.
We begin with a brief review of how to generate CAMs.

Given an image classifier, \textit{e.g.}, ResNet \citep{resnet} and ViT \citep{dosovitskiy2021image}, we denote the last feature maps as $\mathbf{F} \in \mathbb{R}^{d \times hw}$, where $hw$ is the spatial size and $d$ indicates the channel dimension.
The activation map $\mathbf{M}^c$ for class $c$ is generated via weighting the feature maps $\mathbf{F}$ with their contribution to
class $c$:
\begin{equation}
    \mathbf{M}^c = \texttt{ReLU}\left(\sum_{i=1}^d \mathbf{w}_{c,i} \mathbf{F}_{i,:}\right),
\end{equation}
where $\mathbf{w} \in \mathbb{R}^{C \times d}$ is the parameters of the last fully connected layers.
Then, \texttt{min-max} normalization is applied to re-scale $\mathbf{M}^c$ to $[0, 1]$.
$\beta \in (0, 1)$ is used to discriminate foreground regions from background, \textit{i.e.}, when the value of $\mathbf{M}^c$ of a particular pixel is larger then $\beta$, it is regarded as \textit{reliable} predictions as well as pseudo-labels.

At each training step, we randomly sample $B$ images and their corresponding image-level labels, resulting in a batch $\mathcal{B} = \{\mathbf{x}_i, \mathbf{y}_i\}_{i=1}^B$.
Since we only have image-level labels this time, an additional MLP $f_{\mathrm{cls}}$ is introduced to perform image-level classification, and the supervised loss is thus the \textit{multi-label} soft margin loss:
\begin{equation}
    \mathcal{L}_s = \frac{1}{|\mathcal{B}|} \sum_{(\mathbf{x}_i, \mathbf{y}_i) \in \mathcal{B}} \ell_{ce}(f_{\mathrm{cls}}\circ h(\mathbf{x}_i; \theta).
\end{equation}
As for the unsupervised term 
$\mathcal{L}_u$, it is the vanilla cross-entropy loss that leverages $\mathbf{M}_i$ to be supervision signals:
\begin{equation}
\begin{aligned}
    \mathcal{L}_u = \frac{1}{|\mathcal{B}|} \sum_{(\mathbf{x}_i, \mathbf{y}_i) \in \mathcal{B}} \ell_{ce}(f\circ h(\mathbf{x}_i; \theta), \tilde{\mathbf{y}}_i),
\end{aligned}
\end{equation}
where the pixel-level pseudo-label $\tilde{\mathbf{y}}_i$ is generated by a simple threshold strategy following \citep{ru2022learning}:
\begin{equation}
    \tilde{y}_{ic} = \mathbbm{1}[\mathbf{M}_i^c > \beta],
\end{equation}
where $\mathbbm{1}[\cdot]$ is the indicator function.

\subsection{Elaboration of \method}
\label{sec:elaboration}

In label-efficient learning, discarding unreliable pseudo-labels is widely used to prevent performance degradation \citep{pseudoseg, st++, fixmatch, xie2020self}. 
However, such contempt for unreliable pseudo-labels may result in information loss. 
It is obvious that unreliable pseudo-labels can provide information for better discrimination.
For example, the white cross in Fig.~\ref{fig:example}, is typically an unreliable pixel, whose distribution demonstrates its uncertainty to distinguish between class \texttt{person} and class \texttt{motorbike}.
However, this distribution also demonstrates its certainty \textit{not} to discriminate this pixel as class \texttt{car}, class \texttt{train}, class \texttt{bicycle}, and so on.
Such characteristic gives us the main insight of using unreliable pseudo-labels for label-efficient semantic segmentation.

Mathematically, \method aims to recycle unreliable predictions into training by adding an extra term $\mathcal{L}_c$ into Eq.~(\ref{eq:common}).
Therefore, the overall objective becomes to 
\begin{equation}
    \mathcal{L}_{\mathrm{U^{2}PL+}} = \mathcal{L}_s + \lambda_u \mathcal{L}_u + \lambda_c \mathcal{L}_c,
\end{equation}
where $\lambda_c$ is an extra weight to balance its contribution.
Specifically, $\mathcal{L}_c$ is the pixel-level InfoNCE \citep{infonce} loss:
\begin{equation}
\label{eq:contraloss}
\begin{aligned}
    \mathcal{L}_c = &- \frac{1}{C\times M}\sum_{c=0}^{C-1} \sum_{i=1}^M \\
    &\log\left[ \frac{e^{\langle \mathbf{z}_{ci}, \mathbf{z}_{ci}^{+}\rangle / \tau}}{e^{\langle \mathbf{z}_{ci}, \mathbf{z}_{ci}^{+}\rangle / \tau} + \sum_{j=1}^N e^{\langle \mathbf{z}_{ci}, \mathbf{z}_{cij}^{-}\rangle / \tau}} \right],
\end{aligned}
\end{equation}
where $M$ is the total number of anchor pixels, and $\mathbf{z}_{ci}$ denotes the representation of the $i$-th anchor of class $c$.
Each anchor pixel is followed with a positive key and $N$ negative keys, whose representations are $\mathbf{z}_{ci}^{+}$ and $\{\mathbf{z}_{cij}^{-}\}_{j=1}^N$, respectively.
$\mathbf{z}=g\circ h (\mathbf{x})$ is the output of the representation head.
$\langle \cdot, \cdot \rangle$ is the cosine similarity between features from two different pixels, whose range is limited between $-1$ to $1$, hence the need for temperature $\tau$.
We set $M=256$, $N=50$ and $\tau=0.5$ in practice.
How to select (a) anchor pixels (queries) $\mathbf{z}_{ci}$, (b) the positive key $\mathbf{z}_{ci}^+$ for each anchor, and (c) negative keys $\mathbf{z}_{cij}^-$ for each anchor, is different among different label-efficient settings.

Fig.~\ref{fig:pipeline} illustrates how \method conducts contrastive pairs.
Concretely, predictions are first split into reliable ones and unreliable ones by leveraging the pixel-level entropy map.
Reliable predictions (pixels marked in color other than white) are first averaged and then used to update category-wise prototypes, which are served as positive keys for given queries.
Unreliable predictions (pixels marked in white) are served as negative keys regarding unlikely classes and are pushed into category-wise memory banks.

Next, we introduce how to apply \method to label-efficient semantic segmentation as follows.
Specifically, we first introduce how we produce pseudo-labels in Sec.~\ref{sec:pseudo}.
Next, denoising techniques are described in Sec.~\ref{sec:denoising}.
Finally, how to sample contrastive pairs are explored in Sec.~\ref{sec:contra}.
Note that pseudo-labeling and denoising is used only in semi-supervised and domain-adaptive semantic segmentation.

\subsection{Pseudo-Labeling}
\label{sec:pseudo}
To avoid overfitting incorrect pseudo-labels, we utilize entropy of every pixel's probability distribution to filter high-quality pseudo-labels for further supervision.
Specifically, we denote $\mathbf{p}_{ij}\in\mathbb{R}^C$ as the softmax probabilities generated by the segmentation head of the teacher model for the $i$-th unlabeled image at pixel $j$, where $C$ is the number of classes.
Its entropy is computed by:
\begin{equation}
    \mathcal{H}(\mathbf{p}_{ij}) = -\sum_{c=0}^{C-1} p_{ij}(c)\log p_{ij}(c),
\end{equation}
where $p_{ij}(c)$ is the value of $\mathbf{p}_{ij}$ at $c$-th dimension.

Then, we define pixels whose entropy on top $\alpha_t$ as unreliable pseudo-labels at training epoch $t$. 
Such unreliable pseudo-labels are not qualified for supervision.
Therefore, we define the pseudo-label for the $i$-th unlabeled image at pixel $j$ as:
\begin{equation}
\label{eq:pseudo}
    \hat{y}_{ij}^u = \left\{
    \begin{aligned}
        &\arg\max_c p_{ij}(c), &&\mathrm{if}\ \mathcal{H}(\mathbf{p}_{ij}) < \gamma_t, \\
        &\mathrm{ignore}, &&\mathrm{otherwise},
    \end{aligned}
    \right.
\end{equation}
where $\gamma_t$ represents the entropy threshold at $t$-th training step. 
We set $\gamma_t$ as the quantile corresponding to $\alpha_t$, \textit{i.e.}, $\gamma_t$~=~\texttt{np.percentile(H.flatten(),100*(1-$\alpha_t$))}, where \texttt{H} is per-pixel entropy map.
We adopt the following adjustment strategies in the pseudo-labeling process for better performance.

\paragraph{Dynamic Partition Adjustment.} During the training procedure, the pseudo-labels tend to be reliable gradually.
Base on this intuition, we adjust unreliable pixels' proportion $\alpha_t$ with linear strategy every epoch:
\begin{equation}
\label{eq:dpa}
    \alpha_t = \alpha_0 \cdot \left(1 - \frac{t}{\mathrm{total\ epoch}}\right),
\end{equation}
where $\alpha_0$ is the initial proportion and is set to $20\%$, and $t$ is the current training epoch.

\paragraph{Adaptive Weight Adjustment.} After obtaining reliable pseudo-labels, we involve them in the unsupervised loss in Eq.~(\ref{eq:unsloss}). 
The weight $\lambda_u$ for this loss is defined as the reciprocal of the percentage of pixels with entropy smaller than threshold $\gamma_t$ in the current mini-batch multiplied by a base weight $\eta$:
\begin{equation}
\label{eq:awa}
    \lambda_u = \eta \cdot \frac{|\mathcal{B}_u|\times H\times W}{\sum_{i=1}^{|\mathcal{B}_u|} \sum_{j=1}^{H\times W} \mathbbm{1}\left[\hat{y}_{ij}^u \neq \mathrm{ignore}\right]},
\end{equation}
where $\mathbbm{1}(\cdot)$ is the indicator function and $\eta$ is set to $1$.

\subsection{Pseudo-Label Denoising}
\label{sec:denoising}

It is widely known that self-training-based methods often suffer a lot from confirmation bias \citep{arazo2020pseudo}.
Filtering reliable pseudo-labels \citep{st++, pseudoseg} (which has been introduced in Sec.~\ref{sec:pseudo}) and applying strong data augmentation \citep{cutmix, cutout, classmix} are two typical ways to face this issue.
However, considering the domain shift between the labeled source domain and the unlabeled target domain, the model tends to be over-confident \citep{zhang2021prototypical} and there is not enough to simply select pseudo-labels based on their reliability.
To this end, we maintain a prototype for each class and use them to denoise those pseudo-labels.

Let $\mathbf{z}^{\mathrm{proto}}_c$ be the prototype of class $c$, as known as the center of the representation space $\mathcal{P}_c$
\begin{equation}\label{eq:proto}
    \mathbf{z}^{\mathrm{proto}}_c = \frac{1}{|\mathcal{P}_c|} \sum_{\mathbf{z}_c\in\mathcal{P}_c} \mathbf{z}_c,
\end{equation}
and concretely, $\mathcal{P}_c$ contains all labeled pixels belongs to class $c$ and reliable unlabeled pixels predicted to probably belongs to class $c$:
\begin{equation}
    \mathcal{P}_c = \mathcal{P}_c^l \cup \mathcal{P}_c^u,
\end{equation}
where $\mathcal{P}_c^l$ and $\mathcal{P}_c^u$ denote the representation space of labeled pixels and unlabeled pixels respectively:
\begin{equation}
\begin{aligned}
    \mathcal{P}_c^l &= \left\{ \mathbf{z}_{ij} = g\circ h(\mathbf{x}^l_i; \theta_t)_{j} \mid  y_{ij} = c, (\mathbf{x}^l_i, \mathbf{y}_i)\in\mathcal{B}_l \right\}, \\
    \mathcal{P}_c^u &= \left\{ \mathbf{z}_{ij} = g\circ h(\mathbf{x}^u_i; \theta_t)_{j} \mid  \hat{y}_{ij} = c, (\mathbf{x}^u_i, \mathbf{\hat{y}}_i)\in\mathcal{B}_u \right\}, 
\end{aligned}
\end{equation}
where index $i$ means the $i$-th labeled (unlabeled) image, and index $j$ means the $j$-th pixel of that image.

\paragraph{Momentum Prototype.} For the stability of training, all representations are supposed to be consistent \citep{moco}, forwarded by the teacher model hence, and are momentum updated by the centroid in the current mini-batch.
Specifically, for each training step, the prototype of class $c$ is estimated as
\begin{equation}
\label{eq:mopro}
    \mathbf{z}^{\mathrm{proto}}_c \leftarrow m^{\mathrm{proto}}\mathbf{z}^{\prime\mathrm{proto}}_c + (1 - m^{\mathrm{proto}})\mathbf{z}^{\mathrm{proto}}_c,
\end{equation}
where $\mathbf{z}^{\mathrm{proto}}_c$ is the current center defined by Eq.~(\ref{eq:proto}) and $\mathbf{z}^{\prime\mathrm{proto}}_c$ is the prototype for last training step, and $m^{\mathrm{proto}}$ is the momentum coefficient which is set to $0.999$.

\paragraph{Prototypical Denoising.} Let $\mathbf{w}_{i}\in\mathbb{R}^C$ be the weight between $i$-th unlabeled image $\mathbf{x}_i^u$ and each class.
Concretely, we define $w_i(c)$ as the softmax over feature distances to prototypes:
\begin{equation}
    w_i(c) = \frac{\exp\left(-|| f(\mathbf{x}_i^u; \theta_t) - \mathbf{z}^{\mathrm{proto}}_c||\right)}{\sum_{c=1}^C \exp\left(-|| f(\mathbf{x}_i^u; \theta_t) - \mathbf{z}^{\mathrm{proto}}_c||\right)}.
\end{equation}
\noindent Then, we get denoised prediction vector $\mathbf{p}^{*}_{ij}$ for pixel $j$ of $i$-th unlabeled image based on its original prediction vector $\mathbf{p}_{ij}$ and class weight $w_{ic}$:
\begin{equation}
    \mathbf{p}^{*}_{ij} = \mathbf{w}_{i} \odot \mathbf{p}_{ij},
\end{equation}
where $\odot$ denotes the element-wise dot.

Finally, we take $\mathbf{p}^{*}_{ij}$ into Eq.~(\ref{eq:pseudo}) and get denoised pseudo-label $\tilde{y}^u_{ij} = \arg\max_c p^{*}_{ij}$ for pixel $j$ of the $i$-th unlabeled image in Eq.~(\ref{eq:unsloss}).

\subsection{Contrastive Pairs Sampling}
\label{sec:contra}

Formulated in Eq.~(\ref{eq:contraloss}), our \method aims to make sufficient use of unreliable predictions by leveraging an extra contrastive objective that pulls positive pairs $(\mathbf{z}, \mathbf{z}^+)$ together while pushes negatives pairs $(\mathbf{z}, \mathbf{z}^-)$ away.
In the following, we provide a detailed description on conducting contrastive pairs for both settings.

\begin{algorithm}[t]
  \SetAlgoLined

  Initialize $\mathcal{L} \leftarrow 0$\;
  Sample labeled images $\mathcal{B}_l$ and unlabeled images $\mathcal{B}_u$\;
  
  \For{$\mathbf{x}_i \in \mathcal{B}_l\cup\mathcal{B}_u$}{
    Get probabilities: $\mathbf{p}_i \leftarrow f\circ h(\mathbf{x}_i;\theta_t)$\;
    Get representations: $\mathbf{z}_i \leftarrow g\circ h(\mathbf{x}_i;\theta_s)$\;
    
    \For{$c \leftarrow 0$ \KwTo $C-1$}{
      Get anchors $\mathcal{A}_c$ based on Eq.~(\ref{eq:ac}) or Eq.~(\ref{eq:ac_ws})\;
      Sample $M$ anchors: $\mathcal{B}_A \leftarrow $ \texttt{sample} $(\mathcal{A}_c)$\;
      
      Get negatives $\mathcal{N}_c$ based on Eq.~(\ref{eq:negative}) or Eq.~(\ref{eq:negative_ws})\;
      Push $\mathcal{N}_c$ into memory bank $\mathcal{Q}_c$\; 
      Pop oldest ones out of $\mathcal{Q}_c$ if necessary\;
      Sample $N$ negatives: $\mathcal{B}_N \leftarrow $ \texttt{sample} $(\mathcal{Q}_c)$\;
      
      Get $\mathbf{z}^{+}$ based on Eq.~(\ref{eq:positive}) or Eq.~(\ref{eq:proto_ws})\;
      
      $\mathcal{L} \leftarrow \mathcal{L} + \ell(\mathcal{B}_A,\mathcal{B}_N, \mathbf{z}^{+})$ based on Eq.~(\ref{eq:contraloss})\;
    }
  }
  
  \KwOut{contrastive loss $\mathcal{L}_c \leftarrow \frac{1}{|\mathcal{B}|\times C} \mathcal{L}$}
  \caption{Using Unreliable Pseudo-Labels}
  \label{algo:u2pl}
\end{algorithm}

\subsubsection{For SS and DA Settings}

\noindent\textbf{Anchor Pixels (Queries).}
During training, we sample anchor pixels (queries) for each class that appears in the current mini-batch.
We denote the set of features of all labeled candidate anchor pixels for class $c$ as $\mathcal{A}_c^l$, 
\begin{equation}
\label{eq:Al}
    \mathcal{A}_c^l = \left\{
    \mathbf{z}_{ij} \mid y_{ij}=c, p_{ij}(c) > \delta_p
    \right\},
\end{equation}
where $y_{ij}$ is the ground truth for the $j$-th pixel of labeled image $i$, and $\delta_p$ denotes the positive threshold for a particular class and is set to $0.3$ following \citep{reco}. 
$\mathbf{z}_{ij}$ means the representation of the $j$-th pixel of labeled image $i$.
For unlabeled data, counterpart $\mathcal{A}_c^u$ can be computed as:
\begin{equation}
\label{eq:Au}
    \mathcal{A}_c^u = \left\{
    \mathbf{z}_{ij} \mid \tilde{y}_{ij}=c, p_{ij}(c) > \delta_p
    \right\}.
\end{equation}
It is similar to $\mathcal{A}_c^l$, and the only difference is that we use pseudo-label $\hat{y}_{ij}$ based on Eq.~(\ref{eq:pseudo}) rather than the hand-annotated label, which implies that qualified anchor pixels are reliable, \textit{i.e.}, $\mathcal{H}(\mathbf{p}_{ij}) \leq \gamma_t$.
Therefore, for class $c$, the set of all qualified anchors is
\begin{equation}
\label{eq:ac}
    \mathcal{A}_c=\mathcal{A}_c^l\cup\mathcal{A}_c^u.
\end{equation}

\paragraph{Positive Keys.}
The positive sample is the same for all anchors from the same class.
It is the prototype of class $c$ defined in Eq.~(\ref{eq:proto}):
\begin{equation}
\label{eq:positive}
    \mathbf{z}_c^{+} = \mathbf{z}^{\mathrm{proto}}_c.
\end{equation}

\paragraph{Negative Keys.}
We define a binary variable $n_{ij}(c)$ to identify whether the $j$-th pixel of image $i$ is qualified to be negative samples of class $c$.
\begin{equation}
    n_{ij}(c) = \left\{
    \begin{aligned}
        &n_{ij}^l(c), &\mathrm{if\ image\ } i \mathrm{\ is\ labeled}, \\
        &n_{ij}^u(c), &\mathrm{otherwise},
    \end{aligned}
    \right.
\end{equation}
where $n_{ij}^l(c)$ and $n_{ij}^u(c)$ are indicators of whether the $j$-th pixel of labeled and unlabeled image $i$ is qualified to be negative samples of class $c$ respectively.

For $i$-th labeled image, a qualified negative sample for class $c$ should be: (a) not belonging to class $c$; (b) difficult to distinguish between class $c$ and its ground-truth category.
Therefore, we introduce the pixel-level category order $\mathcal{O}_{ij}=\texttt{argsort}(\mathbf{p}_{ij})$. %
Obviously, we have $\mathcal{O}_{ij}(\arg\max\mathbf{p}_{ij})=0$ and $\mathcal{O}_{ij}(\arg\min\mathbf{p}_{ij})=C-1$.
\begin{equation}
\label{eq:nl}
    n_{ij}^l(c) = \mathbbm{1}\left[ y_{ij}\neq c \right] \cdot \mathbbm{1}\left[ 0 \leq \mathcal{O}_{ij}(c) < r_l \right],
\end{equation}
where $r_l$ is the low-rank threshold.
\textcolor{black}{
Here, a small $r_l$ represents an aggressive strategy, which tries to make full use of unreliable predictions but may introduce too much noise.
}
We set to it $3$.
Two indicators reflect (a) and (b), respectively.

For $i$-th unlabeled image, a qualified negative sample for class $c$ should: (a) be unreliable; (b) probably not belongs to class $c$; (c) not belongs to most unlikely classes.
Similarly, we also use $\mathcal{O}_{ij}$ to define $n_{ij}^u(c)$:
\begin{equation}
    n_{ij}^u(c) = \mathbbm{1}\left[ \mathcal{H}(\mathbf{p}_{ij}) > \gamma_t \right] \cdot \mathbbm{1}\left[ r_l \leq \mathcal{O}_{ij}(c) < r_h \right],
\end{equation}
where $r_h$ is the high-rank threshold and is set to $20$.
Finally, the set of negative samples of class $c$ is
\begin{equation}
\label{eq:negative}
    \mathcal{N}_c = \left\{ \mathbf{z}_{ij} \mid n_{ij}(c) = 1 \right\}.
\end{equation}

\paragraph{Category-wise Memory Bank.}
Due to the long tail phenomenon of the dataset, negative candidates in some particular categories are extremely limited in a mini-batch. 
In order to maintain a stable number of negative samples, we use category-wise memory bank $\mathcal{Q}_c$ (FIFO queue) to store the negative samples for class $c$.
\textcolor{black}{
We simply follow MoCo~\citep{moco}, setting the queue size to $65,536$.
A large queue size contributes diverse negative samples while introducing extra computational costs.
}

Finally, the whole process to use unreliable pseudo-labels is shown in Algorithm \ref{algo:u2pl}.
All features of anchors are attached to the gradient, and come from the student hence,
while features of positive and negative samples are from the teacher.

\subsubsection{For the WS Setting}
\label{sec:method_weakly}

The main difference when applying \method to the WS setting is we do not have any pixel-level annotations now, and thus Eqs.~(\ref{eq:Al}) and (\ref{eq:nl}) are invalid.
Also, pseudo-labeling and denoising techniques introduced in Secs.~\ref{sec:pseudo} and \ref{sec:denoising} are \textit{not} necessary here.
This is because applying a simple threshold $\beta \in (0, 1)$ to re-scaled CAMs $\mathbf{M}^c$ to discriminate foreground regions from the background is effective \citep{ru2022learning}.
To this end, the elaboration of \method on weakly supervised semantic segmentation becomes much easier.
Next, how to select (1) anchor pixels, (2) positive keys, and (3) negative keys, are described in detail.

\paragraph{Anchor Pixels (Queries).} 
This time, as segmentation maps of the whole dataset $\mathcal{D}$ are inaccessible, we cannot sample anchors using ground-truth labels as Eq.~(\ref{eq:Al}) does.
Instead, we simply regard those pixels with a value of CAM larger than $\beta$ as candidates:
\begin{equation}
\label{eq:ac_ws}
    \mathcal{A}_c = \{\mathbf{z}_{ij} \mid \mathbf{M}^c_{ij} > \beta\}.
\end{equation}

\paragraph{Positive Keys.}
Similarly, positive keys for each query $\mathbf{q}_c \in \mathcal{A}_c$ are the prototype of class $c$, \textit{i.e.}, $\mathbf{z}_c^+ = \mathbf{z}_c^{\mathrm{proto}}$.
Similarly, the prototypes are momentum updated described in Eq.~(\ref{eq:mopro}), but category-wise centroids of the current mini-batch $\mathbf{z}_c^{\prime\mathrm{proto}}$ are computed by
\begin{equation}
\label{eq:proto_ws}
    \mathbf{z}_c^{\prime\mathrm{proto}} = \frac{1}{|\mathcal{A}_c|} \sum_{\mathbf{z}_c \in \mathcal{A}_c} \mathbf{z}_c.
\end{equation}

\paragraph{Negative Keys.}
Determining negative keys is much easier under the weakly supervised setting because a set of image-level labels is given for each image.
Concretely, given a image $\mathbf{x}_i$, its image-level label is $\mathbf{y}_i \in \{0, 1\}^C$, where $y_i^c = 1$ indicates that class $c$ exists in this image and vise versa.
Therefore, the indicator $n_{ij}(c)$ representing whether pixel $j$ from the $i$-th image is a qualified negative key for class $c$ natually becomes
\begin{equation}
\label{eq:negative_ws}
    n_{ij}(c) = \mathbbm{1}[y_i^c=0]\ \mathrm{or}\ \mathbbm{1}[\mathbf{M}^c_{ij} < \beta],
\end{equation}
where the first term $\mathbbm{1}[y_i^c=0]$ means this image does not contain category $c$ at all, while the second term $\mathbbm{1}[\mathbf{M}^c_{ij} < \beta]$ indicates this prediction is \textit{unreliable} to determine the pixel belongs to category $c$.
The prediction becomes a qualified negative key when \textit{either} of these two conditions is met.


%% file: 4.exp.tex
\section{Experiments}\label{sec:exp}

We conduct experiments on (1) semi-supervised, (2) domain adaptive, and (3) weakly supervised semantic segmentation benchmarks to verify the effectiveness of \method.
The settings, baselines, and quantitative and qualitative results are provided in Secs.~\ref{sec:exp_semi}, \ref{sec:exp_da}, and \ref{sec:exp_weakly}.
Due to limited computational resources, ablation studies are conducted only on semi-supervised benchmarks.

\begin{table}[t]
\centering
\caption{
Comparison with state-of-the-art methods on \textit{classic} \textbf{PASCAL VOC 2012} \texttt{val} set under different partition protocols. 
The labeled images are selected from the original VOC \texttt{train} set, which consists of $1,464$ samples in total.
The fractions denote the percentage of labeled data used for training, followed by the actual number of images.
All the images from SBD are regarded as unlabeled data.
``SupOnly'' stands for supervised training without using any unlabeled data.
\dag\ means we reproduce the approach.
The best performances are highlighted in \textbf{bold} font and the second scores are \underline{underlined}.
}
\label{tab:classic}
\setlength{\tabcolsep}{2pt}
\begin{tabular}{lccccc}
\toprule
Method & 
1/16 (92) & 1/8 (183) & 1/4 (366) & 1/2 (732) & Full (1464) \\
SupOnly & 
45.77 & 54.92 & 65.88 & 71.69 &72.50 \\
\midrule
MT$^\dag$ & 
51.72 & 58.93 & 63.86 & 69.51 & 70.96 \\
CutMix$^\dag$ & 
52.16 & 63.47 & 69.46 & 73.73 & 76.54 \\
PseudoSeg & 
57.60 & 65.50 & 69.14 & 72.41 & 73.23 \\
PC${}^2$Seg & 
57.00 & 66.28 & 69.78 & 73.05 & 74.15 \\
\midrule
U$^\text{2}$PL  & 
\underline{67.98} & 
\underline{69.15} & 
\underline{73.66} & 
\underline{76.16} & 
\underline{79.49} \\
\method  & 
\textbf{69.29} & 
\textbf{73.40} & 
\textbf{75.03} & 
\textbf{77.09} & 
\textbf{79.52} \\
\bottomrule
\end{tabular}
\vspace{-10pt}
\end{table}

\subsection{Experiments on Semi-Supervised Segmentation}
\label{sec:exp_semi}

In this section, we provide experimental results on semi-supervised semantic segmentation benchmarks.
We first describe the experimental setup, including datasets, network structure, evaluation metric, and implementation details. 
Then, we compare with recent methods in Sec.~\ref{sec:sota}, and perform ablation studies on both PASCAL VOC 2012 and Cityscapes in Sec.~\ref{sec:ablation_voc}.
Moreover, in Sec.~\ref{sec:visual_voc}, we provide qualitative segmentation results.

\begin{table*}[t]
\centering
\caption{
Comparison with state-of-the-art methods on \textit{blender} \textbf{PASCAL VOC 2012} \texttt{val} set and \textbf{Cityscapes} \texttt{val} set under different partition protocols. 
For \textit{blender} VOC, all labeled images are selected from the augmented VOC \texttt{train} set, which consists of $10,582$ samples in total.
For Cityscapes, all labeled images are selected from the Cityscapes \texttt{train} set, which contains $2,975$ samples in total.
``SupOnly'' stands for supervised training without using any unlabeled data.
\dag\ means we reproduce the approach.
The best performances are highlighted in \textbf{bold} font and the second scores are \underline{underlined}.
}
\label{tab:blender_city}
\begin{tabular}{lcccc c lcccc}
\toprule
\multirow{2}{*}{Method} & \multicolumn{4}{c}{\textit{Blender} PASCAL VOC 2012} & \quad & \multirow{2}{*}{Method} & \multicolumn{4}{c}{Cityscapes} \\
\cmidrule(lr){2-5}
\cmidrule(lr){8-11}
& 
1/16 (662) & 1/8 (1323) & 1/4 (2646) & 1/2 (5291) &
& &
1/16 (186) & 1/8 (372) & 1/4 (744) & 1/2 (1488) \\
\midrule
SupOnly & 
67.87 & 71.55 & 75.80 & 77.13 & &
SupOnly &
65.74 & 72.53 & 74.43 & 77.83 \\
MT$^\dag$ & 
70.51 & 71.53 & 73.02 & 76.58 & &
MT$^\dag$ & 
69.03 & 72.06 & 74.20 & 78.15 \\
CutMix$^\dag$ & 
71.66 & 75.51 & 77.33 & 78.21 & &
CutMix$^\dag$ & 
67.06 & 71.83 & 76.36 & 78.25 \\
CCT & 
71.86 & 73.68 & 76.51 & 77.40 & &
CCT & 
69.32 & 74.12 & 75.99 & 78.10 \\
GCT &
70.90 & 73.29 & 76.66 & 77.98 & &
GCT & 
66.75 & 72.66 & 76.11 & 78.34 \\
CPS & 
74.48 & 76.44 & 77.68 & 78.64 & &
CPS$^\dag$ & 
69.78 & 74.31 & 74.58 & 76.81 \\
AEL & 
77.20 & 77.57 & 78.06 & 80.29 & &
AEL$^\dag$ & 
74.45 & 75.55 & 77.48 & 79.01 \\
\midrule
U$^\text{2}$PL  &
\underline{77.21} & 
\underline{79.01} & 
\underline{79.30} & 
\underline{80.50} & &
U$^\text{2}$PL  & 
\underline{74.90} & 
\underline{76.48} & 
\underline{78.51} & 
\underline{79.12} \\
\method  &
\textbf{77.23} & \textbf{79.35} & \textbf{80.21} & \textbf{80.78} & &
\method  & 
\textbf{76.09} & \textbf{78.00} & \textbf{79.02} & \textbf{79.62} \\
\bottomrule
\end{tabular}
\end{table*}


\paragraph{Datasets.} 
In \textit{semi-supervised semantic segmentation}, PASCAL VOC 2012 \citep{voc} and Cityscapes \citep{cityscapes} are two widely used datasets for evaluation.
PASCAL VOC 2012 Dataset is a standard semantic segmentation benchmark with 20 semantic classes of objects and 1 class of background.
The training set and the validation set include $1,464$ and $1,449$ images respectively.
Following \citep{ael, st++, cps}, we use SBD \citep{sbd} as the augmented set with $9,118$ additional training images.
Since the SBD dataset is coarsely annotated, \citep{pseudoseg} takes only the standard $1,464$ images as the whole labeled set, while other methods \citep{cps, ael} take all $10,582$ images as candidate labeled data.
Therefore, we evaluate our method on both the \textit{classic} set ($1,464$ candidate labeled images) and the \textit{blender} set ($10,582$ candidate labeled images).
Cityscapes, a dataset designed for urban scene understanding, consists of $2,975$ training images with fine-annotated masks and $500$ validation images.
For each dataset, we compare \method with other methods under $1/2$, $1/4$, $1/8$, and $1/16$ partition protocols.

\paragraph{Network Structure.}
For \textit{SS segmentation}, we use ResNet-101 \citep{resnet} pre-trained on ImageNet-1K \citep{imagenet} as the backbone and DeepLabv3+ \citep{deeplabv3p} as the decoder.
The representation head consists of two \texttt{Conv-BN-ReLU} blocks, where both blocks preserve the feature map resolution, and the first block halves the number of channels, mapping the extracted features into $256$ dimensional representation space.
The architecture of the representation head remains the same in three different settings.
{
\color{black}
The extra representation head introduces an additional $\approx$ 2.8 M parameters, resulting in roughly 10\% more computational overhead compared to the baseline.
}

\paragraph{Evaluation Metric.}
We adopt the mean of Intersection over Union (mIoU) as the metric to evaluate these cropped images.
For the SS segmentation task, all results are measured on the \texttt{val} set on both Cityscapes \citep{cityscapes} and PASCAL VOC 2012 \citep{voc}, where VOC images are center cropped to a fixed resolution while slide window evaluation is used for Cityscapes following common practices \citep{ael, wang2022semi}.

\paragraph{Implementation Details.}
For the training on the \textit{blender} and \textit{classic} PASCAL VOC 2012 dataset, we use stochastic gradient descent (SGD) optimizer with initial learning rate $0.001$,  weight decay as $0.0001$, crop size as $513\times513$, batch size as $16$ and training epochs as $80$.
For the training on the Cityscapes dataset, we also use stochastic gradient descent (SGD) optimizer with an initial learning rate of $0.01$,  weight decay as $0.0005$, crop size as $769\times769$, batch size as $16$ and training epochs as $200$.
In all experiments, the decoder's learning rate is ten times that of the backbone.
We use the poly scheduling to decay the learning rate during the training process: $lr = lr_{\mathrm{base}} \cdot \left(1 - \frac{\mathrm{iter}}{\mathrm{total\ iter}} \right)^{0.9}$.
All SS experiments are conducted with $8$ Tesla V100 GPUs.

\begin{table}[t]
\centering
\caption{
\textbf{Ablation study on using pseudo pixels with different reliability as \textit{negative keys}}.
The reliability is measured by the entropy of pixel-wise prediction (see Sec.~\ref{sec:contra}). 
``U'' denotes \textit{unreliable}, which takes pixels with the \textit{top 20\%} highest entropy scores as negative candidates.
``R'' indicates \textit{reliable}, and means the \textit{bottom 20\%} counterpart. 
We prove this effectiveness under $1/4$ and $1/8$ partition protocols on \textit{blender} PASCAL VOC 2012 \texttt{val} set, and $1/2$ and $1/4$ partition protocols on Cityscapes \texttt{val} set, respectively.
The best performances are highlighted in \textbf{bold} font and the second scores are \underline{underlined}.
}
\label{tab:abalation_reliable}
\setlength{\tabcolsep}{5.5pt}
\begin{tabular}{ccccccc}
\toprule
\multirow{2}{*}{$\mathcal{D}_l$} & \multicolumn{2}{c}{$\mathcal{D}_u$} & \multicolumn{2}{c}{\textit{Blender} VOC} & \multicolumn{2}{c}{Cityscapes} \\
\cmidrule(lr){2-3} 
\cmidrule(lr){4-5} 
\cmidrule(lr){6-7}
& R & U & 1/8 (1323) & 1/4 (2646) & 1/4 (744) & 1/2 (1488) \\
\midrule
\checkmark & & & 74.29 & 75.76 & 74.39 & 78.10 \\
 & \checkmark & & 75.15 & 76.48 & 74.60 & 77.93 \\
 & & \checkmark & \underline{78.37} & \underline{79.01} & \underline{77.19} & \underline{78.16} \\
\checkmark & \checkmark & & 77.30 & 77.35 & 75.16 & 77.19 \\
\checkmark & & \checkmark & \textbf{79.35} & \textbf{80.21} & \textbf{79.02} & \textbf{79.62} \\
\checkmark & \checkmark & \checkmark & 77.40 & 77.57 & 74.51 & 76.96 \\
\bottomrule
\end{tabular}
\vspace{-10pt}
\end{table}

\subsubsection{Comparison with State-of-the-Art Alternatives}
\label{sec:sota}

We compare our method with following recent representative semi-supervised semantic segmentation methods: Mean Teacher (MT) \citep{meanteacher}, CCT \citep{cct},  GCT \citep{gct}, PseudoSeg \citep{pseudoseg}, CutMix \citep{french2019semi}, CPS \citep{cps}, PC${}^2$Seg \citep{pc2seg}, AEL \citep{ael}. 
We re-implement MT \citep{meanteacher}, CutMix \citep{cutmix} for a fair comparison.
For Cityscapes \citep{cityscapes}, we also reproduce CPS \citep{cps} and AEL \citep{ael}.
All results are equipped with the same network architecture (DeepLabv3+ as decoder and ResNet-101 as encoder).
It is important to note the \textit{classic} PASCAL VOC 2012 Dataset and \textit{blender} PASCAL VOC 2012 Dataset only differ in the training set. 
Their validation set is the same with $1,449$ images.

\paragraph{Results on \textit{classic} PASCAL VOC 2012 Dataset.}
Tab.~\ref{tab:classic} compares our method with the other state-of-the-art methods on $\textit{classic}$ PASCAL VOC 2012 Dataset.
\method outperforms the supervised baseline by $+23.52\%$, $+18.48\%$, $+9.15\%$, $+5.40\%$, and $+7.02\%$ under $1/16$, $1/8$, $1/4$, $1/2$, and ``full'' partition protocols respectively, indicating that semi-supervised learning fully mines the inherent information of unlabeled images.
When comparing to state-of-the-arts, our \method outperforms PC${}^2$Seg under all partition protocols by $+12.29\%$, $+7.12\%$, $+5.25\%$, $+4.04\%$, and $+5.37\%$ respectively.
Note that when labeled data is extremely limited, \textit{e.g.}, when we only have $92$ labeled data, our \method outperforms previous methods by a large margin ($+12.29\%$ under $1/16$ split for classic PASCAL VOC 2012),
proofing the efficiency of using unreliable pseudo-labels.
Furthermore, by introducing extra strategies, \method can outperform its previous version U$^\text{2}$PL by $+1.31\%$, $+4.25\%$, $+1.37\%$, $+0.93\%$, and $+0.03\%$ respectively.
The fewer labeled images we have, the larger improvement it brings, indicating that \method is more capable of dealing with training noise than U$^\text{2}$PL.

\paragraph{Results on \textit{blender} PASCAL VOC 2012 Dataset.} 
Tab.~\ref{tab:blender_city} shows the comparison results on \textit{blender} PASCAL VOC 2012 Dataset. 
Our method \method outperforms all the other methods under various partition protocols. 
Compared with the supervised baseline,  \method achieves improvements of $+9.36\%$, $+7.80\%$, $+4.41\%$ and $+3.65\%$ under $1/16$, $1/8$, $1/4$ and $1/2$ partition protocols respectively.
Compared with the existing state-of-the-art methods, \method surpasses them under all partition protocols.
Especially under $1/8$ protocol and $1/4$ protocol, \method outperforms AEL by $+1.78\%$ and $+2.15\%$.
When comparing \method to its previous version U$^\text{2}$PL, it brings improvements of $+0.02\%$, $+0.34\%$, $+0.91\%$, and $+0.28\%$ respectively.
Compared to the classic VOC counterpart, it seems that \method may struggle to bring significant improvements when we access adequate labeled images (\textit{i.e.}, the number of labeled data is extended from $1,464$ to $10,582$).

\begin{table}[t]
\centering
\caption{
\textbf{Ablation study on the effectiveness of various components in our \method}, including unsupervised loss $\mathcal{L}_u$, contrastive loss $\mathcal{L}_c$, category-wise memory bank $\mathcal{Q}_c$, Prototypical Denoising (PD), Momentum Prototype (MP), Symmetric Cross-Entropy (SCE) for unlabeled images, Dynamic Partition Adjustment (DPA), Probability Rank Threshold (PRT), and using unreliable pseudo-labels in contrastive learning (Un).
}
\label{tab:abalation_component}
\begin{tabular}{ccccccccc}
\toprule
$\mathcal{L}_c$ & $\mathcal{Q}_c$ & PD & MP & SCE & DPA & PRT & Un & 1/4 (2646) \\
\midrule
& & & & & & & & 77.33      \\
\checkmark  & & & & & & & & 77.08      \\
\checkmark  & \checkmark  & & & & & \checkmark   & \checkmark   & 78.49      \\
\checkmark  & \checkmark  & & & & \checkmark   & & \checkmark   & 79.07      \\
\checkmark  & \checkmark  & & & & \checkmark   & \checkmark   & & 77.57      \\
\checkmark  & \checkmark  & & & & \checkmark   & \checkmark   & \checkmark & 79.30      \\
& & \checkmark & & & & & & 77.93 \\
& & & & \checkmark   & & & & 77.69 \\
& & \checkmark & & \checkmark   & & & & 78.01 \\
& & \checkmark & \checkmark & \checkmark   & & & & 78.24 \\
\checkmark & \checkmark & \checkmark & \checkmark & \checkmark & \checkmark & \checkmark & \checkmark & \textbf{80.21} \\
\bottomrule
\end{tabular}
\end{table}

\begin{table}[t]
\centering
\caption{
\textbf{Ablation study on prototypical denoising (PD)} on \textit{blender} PASCAL VOC 2012 \texttt{val} set under different partition protocols.
}
\label{tab:abla_denoise}
\setlength{\tabcolsep}{4pt}
\begin{tabular}{lcccc}
\toprule
 & 
1/16 (662) & 1/8 (1323) & 1/4 (2646) & 1/2 (5291) \\
\midrule
\method (w/o PD) &
77.09 & 79.10 & 79.48 & 80.43 \\
\method (w/ PD) &
\textbf{77.23} & \textbf{79.35} & \textbf{80.21} & \textbf{80.78} \\
\bottomrule
\end{tabular}
\end{table}

\begin{table}[t]
    \centering
    \caption{
        \textbf{Ablation study on $\beta$} introduced in Eq.~(\ref{eq:mopro}) on \textit{blender} PASCAL VOC \texttt{val} set and Cityscapes \texttt{val} set.
    }
    \label{tab:abla_mopro}
    \setlength{\tabcolsep}{8pt}
    \begin{tabular}{ccccc}
    \toprule
    \multirow{2}{*}{$\beta$} & \multicolumn{2}{c}{\textit{Blender} VOC} & \multicolumn{2}{c}{Cityscapes} \\
    \cmidrule(lr){2-3}
    \cmidrule(lr){4-5}
    & 1/8 (1323) & 1/4 (2646) & 1/4 (744) & 1/2 (1488) \\
    \midrule
    0.9 & 79.07 & 79.29 & 77.81 & 78.51 \\
    0.99 & 78.91 & 79.37 & 77.93 & 78.54 \\
    0.999 & \textbf{79.35} & \textbf{80.21} & \textbf{79.02} & \textbf{79.62} \\
    0.9999 & 79.18 & 79.91 & 78.82 & 79.11 \\
    \bottomrule
    \end{tabular}
\end{table}

\begin{table}[t]
\centering
\caption{
\textbf{Ablation study on $(\xi_1, \xi_2)$} on \textit{blender} PASCAL VOC 2012 \texttt{val} set and Cityscapes \texttt{val} set.
}
\label{tab:abla_sce}
\begin{tabular}{cccccc}
\toprule
\multirow{2}{*}{$\xi_1$} & \multirow{2}{*}{$\xi_2$} & \multicolumn{2}{c}{\textit{Blender} VOC} & \multicolumn{2}{c}{Cityscapes} \\
\cmidrule(lr){3-4}
\cmidrule(lr){5-6}
& & 1/8 (1323) & 1/4 (2646) & 1/8 (372) & 1/4 (744) \\
\midrule
0 & 1 & 77.76 & 78.63 & 76.71 & 78.02 \\
1 & 0 & 79.06 & 79.68 & 77.03 & 78.53 \\
0.1 & 1 & 79.11 & 79.83 & 77.43 & 78.72 \\
1 & 0.5 & \textbf{79.35} & \textbf{80.21} & \textbf{78.00} & \textbf{79.02} \\
0.5 & 0.5 & 79.33 & 80.20 & 77.79 & 78.81 \\
\bottomrule
\end{tabular}
\vspace{-10pt}
\end{table}

\paragraph{Results on Cityscapes Dataset.}
Tab.~\ref{tab:blender_city} illustrates the comparison results on the Cityscapes \texttt{val} set. 
\method improves the supervised only baseline by $+10.35\%$, $+5.47\%$, $+4.59\%$ and $+1.79\%$ under $1/16$, $1/8$, $1/4$, and $1/2$ partition protocols, and outperforms existing state-of-the-art methods by a notable margin.
In particular, \method outperforms AEL by $+1.64\%$, $+2.45\%$, $+1.54\%$ and $+0.61\%$ under $1/16$, $1/8$, $1/4$, and $1/2$ partition protocols
Compared to its previous version U$^\text{2}$PL, \method achieves improvements of $+1.19\%$, $+1.52\%$, $+0.51\%$, and $+0.50\%$ respectively.

\subsubsection{Ablation Studies}
\label{sec:ablation_voc}

In this section, we first design experiments in Tab.~\ref{tab:abalation_reliable} to validate our main insight: using unreliable pseudo-labels is significant for semi-supervised semantic segmentation.
Next, we ablate each component of our proposed \method in Tab.~\ref{tab:abalation_component}, including using contrastive learning ($\mathcal{L}_c$), applying a memory bank to store abundant negative samples ($\mathcal{Q}_c$), prototype-based pseudo-labels denoising (PD), momentum prototype (MP), using symmetric cross-entropy for unlabeled images (SCE), dynamic partition adjustment (DPA), probability rank threshold (PRT), and only regard unreliable pseudo-labels as negative samples (Un).
In Tab.~\ref{tab:abla_denoise}, we evaluate the performance of prototypical denoising under different partition protocols on both PASCAL VOC 2012 and Cityscapes.
Finally, we ablate the hyper-parameter $\beta$ for momentum prototype, $(\xi_1, \xi_2)$ for symmetric cross-entropy loss, $(r_l, r_h)$ for probability rank threshold, initial reliable-unreliable partition $\alpha_0$, base learning rate $lr_{\mathrm{base}}$, and temperature $\tau$, respectively.

\paragraph{Effectiveness of Using Unreliable Pseudo-Labels.} 
To prove our core insight, \textit{i.e.}, using unreliable pseudo-labels promotes semi-supervised semantic segmentation,
we conduct experiments about selecting negative candidates (which is described in Sec.~\ref{sec:contra}) with different reliability, \textit{i.e.}, whether to regard only unreliable pseudo-labels to be negative samples of a particular query.

Tab.~\ref{tab:abalation_reliable} demonstrates the mIoU results on PASCAL VOC 2012 \texttt{val} set and Cityscapes \texttt{val} set, respectively.
Containing ``U'' in negative keys outperforms other options, proving using unreliable pseudo-labels does help, and our \method fully mines the information of all pixels, especially those unreliable ones.
Note that ``R'' in Tab.~\ref{tab:abalation_reliable} indicates negative keys in memory banks are reliable.
Containing both ``U'' and ``R'' means that all features are stored in memory banks without filtering.
It is worth noticing that containing features from labeled set $\mathcal{D}_l$ brings marginal improvements, but significant improvements are brought because of the introduction of unreliable predictions, \textit{i.e.}, ``U'' in Tab.~\ref{tab:abalation_reliable}.

\paragraph{Effectiveness of Components.}
We conduct experiments in Tab.~\ref{tab:abalation_component} to ablate each component of \method step by step. 
For a fair comparison, all the ablations are under 1/4 partition protocol on the blender PASCAL VOC 2012 Dataset. 

Above all, we use no $\mathcal{L}_{c}$ trained model as our baseline, achieving mIoU of $77.33\%$ (CutMix in Tab.~\ref{tab:blender_city}).
We first ablate components in contrastive learning, including $\mathcal{Q}_c$, DPA, PRT, and Un.
Simply adding vanilla $\mathcal{L}_{c}$ even contributes to the performance degradation by $-0.27\%$.
Category-wise memory bank $\mathcal{Q}_{c}$, along with PRT and high entropy filtering brings an improvement by $+1.41\%$ to vanilla $\mathcal{L}_c$.
Dynamic Partition Adjustment (DPA) together with high entropy filtering, brings an improvement by $+1.99\%$ to vanilla $\mathcal{L}_c$.
Note that DPA is a linear adjustment without tuning (refer to Eq.~(\ref{eq:dpa})), which is simple yet efficient.
For Probability Rank Threshold (PRT) component, we set the corresponding parameter according to Tab.~\ref{tab:abalation_probability_rank}. 
Without high entropy filtering, the improvement decreased significantly to $+0.49\%$.

Then, we ablate components in noisy label learning and its denoising strategies, including PD, MP, and SCE.
Introducing Prototypical Denoising (PD) to pseudo-labels improves the performance by $+0.60\%$ to CutMix.
Using Symmetric Corss-Entropy (SCE) in computing unsupervised loss, improves the performance by $+0.36\%$ to CutMix.
Adding them together brings an improvement by $+0.68\%$.
Adopting an extra Momentum Prototype (MP) on the basis of the above two techniques brings an improvement by $+0.91\%$.

Finally, when adding all the contributions together, our method achieves state-of-the-art results under $1/4$ partition protocol with mIoU of $80.21\%$.
Following this result, we apply these components and corresponding parameters in all experiments on Tab.~\ref{tab:classic} and Tab.~\ref{tab:blender_city}.

\begin{table}[t]
\centering
\caption{
\textbf{Ablation study on PRT} on PASCAL VOC 2012 \texttt{val} set and Cityscapes \texttt{val} set.
}
\label{tab:abalation_probability_rank}
\setlength{\tabcolsep}{7pt}
\begin{tabular}{cccccc}
\toprule
\multirow{2}{*}{$r_l$} & \multirow{2}{*}{$r_h$} & \multicolumn{2}{c}{\textit{Blender} VOC} & \multicolumn{2}{c}{Cityscapes} \\
\cmidrule(lr){3-4}
\cmidrule(lr){5-6}
& & 1/8 (1323) & 1/4 (2646) & 1/8 (372) & 1/4 (744) \\
\midrule
1 & 3  & 78.57  & 79.03 & 73.44 & 77.27 \\
1 & 20 & 78.64  & 79.07 & 75.03 & 78.04 \\
3 & 10 & 78.27  & 78.91 & 76.12 & 78.01 \\
3 & 20 & \textbf{79.35} & \textbf{80.21} & \textbf{78.00} & \textbf{79.02} \\
10 & 20 & 78.62 & 78.94 & 75.33 & 77.18 \\
\bottomrule
\end{tabular}
\end{table}

\begin{table}[t]
\centering
\caption{
\textbf{Ablation study on $\alpha_0$ in Eq.~(\ref{eq:dpa})} on \textit{blender} PASCAL VOC 2012 \texttt{val} set and Cityscapes \texttt{val} set, which controls the initial proportion between reliable and unreliable pixels.
}
\setlength{\tabcolsep}{9pt}
\label{tab:abalation_alpha}
\begin{tabular}{ccccc}
\toprule
\multirow{2}{*}{$\alpha_0$} & \multicolumn{2}{c}{\textit{Blender} VOC} & \multicolumn{2}{c}{Cityscapes} \\
\cmidrule(lr){2-3}
\cmidrule(lr){4-5}
& 1/8 (1323) & 1/4 (2646) & 1/8 (372) & 1/4 (744) \\
\midrule
40\% & 76.77 & 76.92 & 75.07 & 77.20 \\
30\% & 77.34 & 76.38 & 75.93 & 78.08 \\
20\% & \textbf{79.35} & \textbf{80.21} & \textbf{78.00} & \textbf{79.02} \\
10\% & 77.80 & 77.95 & 74.63 & 78.40 \\
\bottomrule
\end{tabular}
\end{table}

\begin{table}[t]
\centering
\caption{
\textbf{Ablation study on base learning rate and temperature} under 1/4 partition protocol (2646) on \textit{blender} VOC PASCAL 2012 Dataset.
}
\setlength{\tabcolsep}{9pt}
\label{tab:baselr}
\begin{tabular}{cccccc}
\toprule
$lr_{\mathrm{base}}$& 
$10^{-1}$ & $10^{-2}$ & $10^{-3}$ & $10^{-4}$ &  $10^{-5}$ \\
\midrule
mIoU 
& 3.49 & 77.82 &\textbf{80.21} &74.58  &65.69 \\
\midrule
$\tau$& 
$10$ & $1$ & $0.5$ & $0.1$ &  $0.01$ \\
\midrule
mIoU & 
78.88 & 78.91 &\textbf{80.21} &79.22  &78.78 \\
\bottomrule
\end{tabular}
\vspace{-10pt}
\end{table}

\paragraph{Effectiveness of Prototypical Denoising.}
\label{sec:abla_denoise}
To further see the impact of prototypical denoising (PD), we conduct experiments in Tab.~\ref{tab:abla_denoise} to find out how PD affects the performance under different partition protocols of \textit{blender} PASCAL VOC 2012.
From Tab.~\ref{tab:abla_denoise}, we can tell that PD brings improvements of $+0.14\%$, $+0.25\%$, $+0.73\%$, and $+0.35\%$ under $1/16$, $1/8$, $1/4$, and $1/2$ partition protocols respectively,
indicating that PD manages to enhance the quality of pseudo-labels by weighted averaging predictions from similar feature representations.

\begin{table}[t]
    \centering
    \caption{
    Semi-supervised semantic segmentation on \textbf{Cityscapes} with \textit{transformer-based} network architecture
    We adopt DAFormer~\citep{hoyer2022daformer} with MiT-B5~\citep{xie2021segformer}.
    All methods are trained with 40k iterations for efficient evaluation.
    The input resolution is $512\times512$.
    %
    %
    \textit{\method manages to bring significant improvements when using transformer-based models}.
    }
    \label{tab:daformer_ss}
    \begin{tabular}{l cccc}
    \toprule
    Method & 1/16 (186) & 1/8 (372) & 1/4 (744) & 1/2 (1488) \\
    \midrule
    SupOnly & 63.11 & 68.84 & 70.83 & 74.56 \\
    CutMix & 67.84 & 71.29 & 72.37 & 75.46 \\
    \method & \textbf{72.45} & \textbf{75.03} & \textbf{75.70} & \textbf{76.75} \\
    \bottomrule
    \end{tabular}
\end{table}

\begin{table}[t]
    \centering
    \caption{
    Experiments on semi-supervised image classification.
    We tested our \method on \textbf{CIFAR-100}~\citep{cifar}.
    }
    \label{tab:fixmatch}
    \setlength{\tabcolsep}{5pt}
    \begin{tabular}{l cccc}
    \toprule
    Method & 1/125 (400) & 1/20 (2500) & 1/5 (10000) \\
    \midrule
    FixMatch & 51.15 & 71.71 & 77.40 \\
    \method (w/ FixMatch) & \textbf{57.33} & \textbf{73.07} & \textbf{79.02} \\
    FreeMatch & 62.02 & 73.53 & 78.32 \\
    \method (w/ FreeMatch) & \textbf{63.74} & \textbf{74.01} & \textbf{79.18} \\
    \bottomrule
    \end{tabular}
    \vspace{-10pt}
\end{table}

\paragraph{Momentum Coefficient $\beta$ for Updating Prototypes.}
We conduct experiments in Tab.~\ref{tab:abla_mopro} to find out the best $\beta$ for momentum prototype (MP) on $1/8$ and $1/4$ partition protocols on \textit{blender} PASCAL VOC 2012, and $1/4$ and $1/2$ partition protocols on Cityscapes, respectively
$\beta = 0.999$ yields slightly better than other settings, indicating that our \method is quite robust against different $\beta$.

\paragraph{Weights in Symmetric Cross-Entropy Loss $\xi_1$ and $\xi_2$.}
To find the optimal $\xi_1$ and $\xi_2$ for symmetric cross-entropy (SCE) loss, we conduct experiments in Tab.~\ref{tab:abla_sce} under $1/8$ and $1/4$ partition protocols on both \textit{blender} PASCAL VOC 2012 and Cityscapes.
When $\xi=1$ and $\xi_2=0.1$, it achieves the best performance, better than standard cross-entropy loss (\textit{i.e.}, when $\xi_1=1$ and $\xi_2=0$) by $+0.29\%$ and $+0.53\%$ on $1/8$ and $1/4$ partition protocols respectively.
Note that the performances drop heavily when using only the reserve CE loss (\textit{i.e.}, $\xi_1=0$ and $\xi_2=1$).

\paragraph{Probability Rank Thresholds $r_l$ and $r_h$.}
Sec.~\ref{sec:contra} proposes to use a probability rank threshold to balance informativeness and confusion caused by unreliable pixels.
Tab.~\ref{tab:abalation_probability_rank} provides a verification that such balance promotes performance.
$r_l = 3$ and $r_h = 20$ outperform other options by a large margin.
When  $r_l = 1$, false negative candidates would not be filtered out, causing the intra-class features of pixels incorrectly distinguished by $\mathcal{L}_{c}$. 
When  $r_l = 10$, negative candidates tend to become irrelevant with corresponding anchor pixels in semantics, making such discrimination less informative.

\paragraph{Initial Reliable-Unreliable Partition $\alpha_0$.} 
Tab.~\ref{tab:abalation_alpha} studies the impact of different initial reliable-unreliable partition $\alpha_0$.
$\alpha_0$ has a certain impact on performance. We find $\alpha_0 = 20\%$ achieves the best performance.
Small $\alpha_0$ will introduce incorrect pseudo labels for supervision, and large $\alpha_0$ will make the information of some high-confidence samples underutilized.

\paragraph{Base Learning Rate $lr_{\mathrm{base}}$.}
The impact of the base learning rate is shown in Tab.~\ref{tab:baselr}. 
Results are based on the \textit{blender} VOC PASCAL 2012 dataset. 
We find that 0.001 outperforms other alternatives.

\paragraph{Temperature $\tau$.}
Tab.~\ref{tab:baselr} gives a study on the effect of temperature $\tau$.
Temperature $\tau$ plays an important role to adjust the importance of hard samples
When $\tau=0.5$, our \method achieves the best results.
Too large or too small of $\tau$ will have an adverse effect on overall performance.

\begin{figure}[t]
    \centering
    \includegraphics[width=1\linewidth]{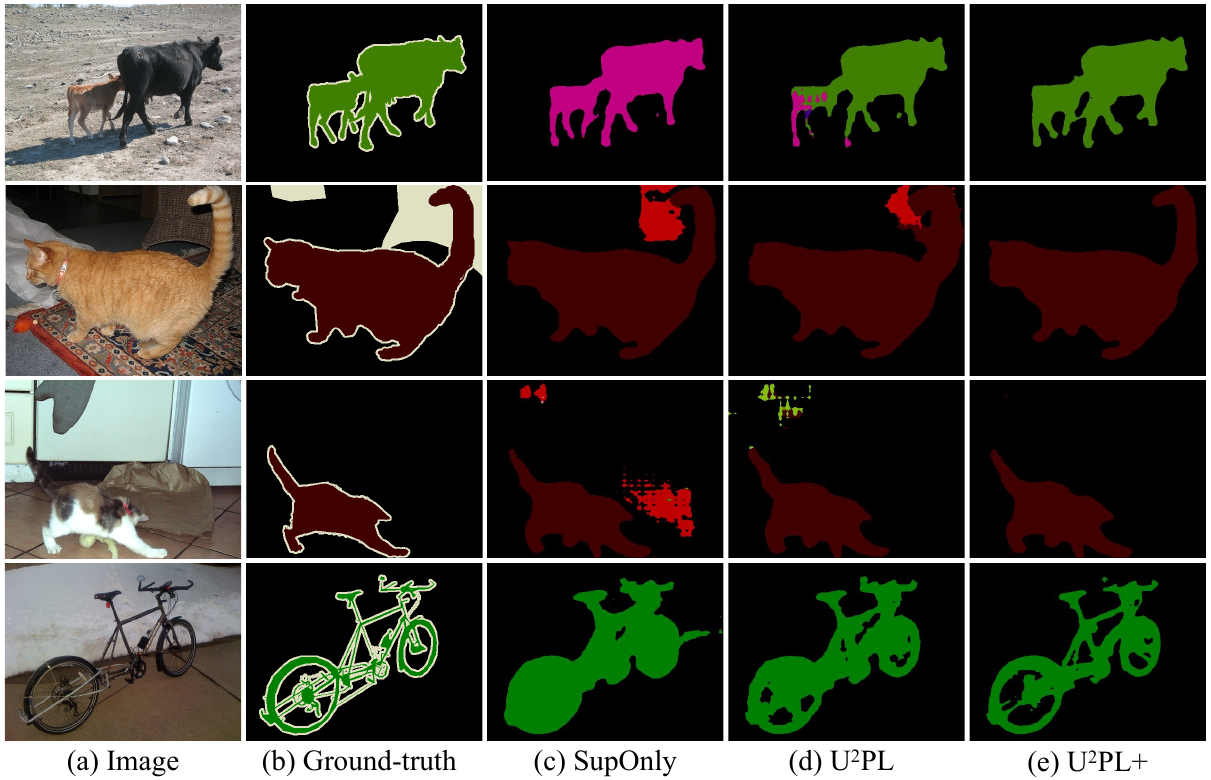}
    \vspace{-13pt}
    \caption{
    Qualitative results on \textbf{PASCAL VOC 2012} \texttt{val} set.
    All models are trained under the $1/4$ partition protocol of \textit{blender} set, which contains $2,646$ labeled images and $7,396$ unlabeled images.
    (a) Input images. 
    (b) Labels for the corresponding image. 
    (c) \textit{Only} labeled images are used for training without any unlabeled data.
    (d) Predictions from our conference version U$^\text{2}$PL.
    (e) Predictions from \method.
    %
    }
    \label{fig:visual}
    \vspace{-10pt}
\end{figure}

{
\color{black}

\paragraph{Extend \method to Transformer-based Models.}
To further verify the generalization of \method towards different network architectures, We apply the network architecture of DAFormer~\citep{hoyer2022daformer} on semi-supervised benchmarks in Tab.~\ref{tab:daformer_ss}.
We train models for 40k iterations instead of 200 epochs for efficient evaluation.
\method manages to bring significant improvements when using transformer-based models.

\paragraph{Extend \method to Semi-Supervised Image Classification.}
To further demonstrate the generalization of \method, we extend our \method to semi-supervised image classification.
We implement our \method on the basis of FixMatch \citep{fixmatch} and FreeMatch \citep{wang2023freematch}, and almost no modifications are needed when adapting our \method from segmentation to classification.
Similarly, we tend to make full use of unreliable \textit{image-level} pseudo labels under the pipeline of \method.
Our implementation is based on USB \citep{usb2022}.
From Tab.~\ref{tab:fixmatch}, we can tell that incorporating \method brings significant improvements.

}

\begin{figure*}[t]
    \centering
    \includegraphics[width=1\linewidth]{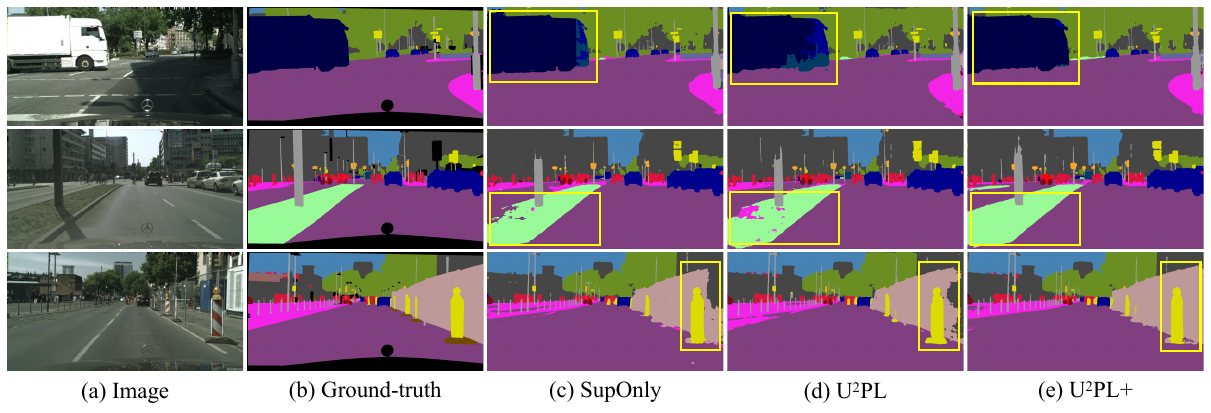}
    \vspace{-15pt}
    \caption{
    Qualitative results on \textbf{Cityscapes} \texttt{val} set.
    All models are trained under the $1/2$ partition protocol, which contains $1,488$ labeled images and $1,487$ unlabeled images.
    (a) Input images. 
    (b) Hand-annotated labels for the corresponding image. 
    (c) \textit{Only} labeled images are used for training.
    (d) Predictions from our conference version, \textit{i.e.}, U$^\text{2}$PL \citep{wang2022semi}.
    (e) Predictions from \method.
    Yellow rectangles highlight the promotion by adequately using unreliable pseudo-labels.
    }
    \label{fig:visual_city}
    \vspace{-10pt}
\end{figure*}

\subsubsection{Qualitative Results}
\label{sec:visual_voc}

Fig.~\ref{fig:visual} and Fig.~\ref{fig:visual_city} show the results of different methods on the PASCAL VOC 2012 \texttt{val} set and Cityscapes \texttt{val} set, respectively.
Benefiting from using unreliable pseudo-labels, \method outperforms the supervised baseline, and \method is able to generate a more accurate segmentation map.

Furthermore, through visualizing the segmentation results, we find that our method achieves much better performance on those ambiguous regions (\textit{e.g.}, the border between different objects).
Such visual difference proves that our method finally makes the reliability of unreliable prediction labels stronger.




\begin{table*}[t]
\centering
\caption{
Comparison with state-of-the-art methods for DA. 
The results are averaged over 3 random seeds.
Note that for SYNTHIA $\to$ Cityscapes, we only compute the mIoU over 16 classes (mIoU$^{16}$).
The top performance is highlighted in \textbf{bold} font and the second score is \underline{underlined}.
}
\setlength{\tabcolsep}{3.5pt}
\label{tab:DA}
\begin{tabular}{l | ccccccccccccccccccc | c}
\toprule
\multicolumn{21}{c}{GTA5 $\to$ Cityscapes} \\
\midrule
Method &
\rotatebox{90}{Road} & \rotatebox{90}{S.walk} & \rotatebox{90}{Build.} & \rotatebox{90}{Wall} & \rotatebox{90}{Fence} & \rotatebox{90}{Pole} & \rotatebox{90}{T.light} & \rotatebox{90}{Sign} & \rotatebox{90}{Veget.} & \rotatebox{90}{Terrain} & \rotatebox{90}{Sky} & \rotatebox{90}{Person} & \rotatebox{90}{Rider} & \rotatebox{90}{Car} & \rotatebox{90}{Truck} & \rotatebox{90}{Bus} & \rotatebox{90}{Train} & \rotatebox{90}{M.bike} & \rotatebox{90}{Bike} & mIoU \\
\midrule
AdaptSeg
& 86.5 & 25.9    & 79.8     & 22.1 & 20.0  & 23.6 & 33.1  & 21.8 & 81.8       & 25.9    & 75.9 & 57.3   & 26.2  & 76.3 & 29.8  & 32.1 & 7.2   & 29.5      & 32.5 & 41.4 \\
CyCADA  
& 86.7 & 35.6    & 80.1     & 19.8 & 17.5  & 38.0 & 39.9  & 41.5 & 82.7       & 27.9    & 73.6 & 64.9   & 19.0  & 65.0 & 12.0  & 28.6 & 4.5   & 31.1      & 42.0 & 42.7 \\
ADVENT 
& 89.4 & 33.1    & 81.0     & 26.6 & 26.8  & 27.2 & 33.5  & 24.7 & 83.9       & 36.7    & 78.8 & 58.7   & 30.5  & 84.8 & 38.5  & 44.5 & 1.7   & 31.6      & 32.4 & 45.5 \\
CBST
& 91.8 & 53.5    & 80.5     & 32.7 & 21.0  & 34.0 & 28.9  & 20.4 & 83.9       & 34.2    & 80.9 & 53.1   & 24.0  & 82.7 & 30.3  & 35.9 & 16.0  & 25.9      & 42.8 & 45.9 \\
FADA
& 92.5 & 47.5    & 85.1     & 37.6 & 32.8  & 33.4 & 33.8  & 18.4 & 85.3       & 37.7    & 83.5 & 63.2   & 39.7  & 87.5 & 32.9  & 47.8 & 1.6   & 34.9      & 39.5 & 49.2 \\
CAG\_DA
& 90.4 & 51.6    & 83.8     & 34.2 & 27.8  & 38.4 & 25.3  & 48.4 & 85.4       & 38.2    & 78.1 & 58.6   & 34.6  & 84.7 & 21.9  & 42.7 & 41.1 & 29.3      & 37.2 & 50.2 \\
FDA
& 92.5 & 53.3    & 82.4     & 26.5 & 27.6  & 36.4 & 40.6  & 38.9 & 82.3       & 39.8    & 78.0 & 62.6   & 34.4  & 84.9 & 34.1  & 53.1 & 16.9  & 27.7      & 46.4 & 50.5 \\
PIT
& 87.5 & 43.4    & 78.8     & 31.2 & 30.2  & 36.3 & 39.3  & 42.0 & 79.2       & 37.1    & 79.3 & 65.4   & 37.5  & 83.2 & 46.0  & 45.6 & 25.7  & 23.5      & 49.9 & 50.6 \\
IAST   
& \underline{93.8} & 57.8   & 85.1     & 39.5 & 26.7  & 26.2 & 43.1  & 34.7 & 84.9       & 32.9    & 88.0 & 62.6   & 29.0  & 87.3 & 39.2  & 49.6 & 23.2  & 34.7      & 39.6 & 51.5 \\
ProDA
& 91.5 & 52.4    & 82.9     & 42.0 & 35.7  & 40.0 & 44.4  & 43.3 & 87.0       & 43.8   & 79.5 & 66.5   & 31.4  & 86.7 & 41.1  & 52.5 & 0.0   & 45.4       & 53.8 & 53.7 \\  
%
DAFormer
& \textbf{95.7} & \underline{70.2}    & \underline{89.4}     & \textbf{53.5} & \textbf{48.1}  & \textbf{49.6} & \underline{55.8}  & \textbf{59.4} & \underline{89.9}       & \underline{47.9}   & \underline{92.5} & \underline{72.2}   & \textbf{44.7}  & \underline{92.3} & \underline{74.5}  & \textbf{78.2} & \underline{65.1}   & \underline{55.9}      & \underline{61.8} & \underline{68.3} \\ 
\midrule
\method 
& \textbf{95.7} & \textbf{70.9}    & \textbf{89.7}     & \underline{53.3} & \underline{46.3}  & \underline{47.0} & \textbf{59.1}  & \underline{58.3} & \textbf{90.4}       & \textbf{48.6}   & \textbf{92.9} & \textbf{73.2}   & \underline{44.6}  & \textbf{92.6} & \textbf{77.5}  & \underline{77.8} & \textbf{76.7}   & \textbf{59.3}      & \textbf{67.5} & \textbf{69.6} \\ 
\midrule
\multicolumn{21}{c}{SYNTHIA $\to$ Cityscapes} \\
\midrule
Method &
\rotatebox{90}{Road} & \rotatebox{90}{S.walk} & \rotatebox{90}{Build.} & \rotatebox{90}{Wall} & \rotatebox{90}{Fence} & \rotatebox{90}{Pole} & \rotatebox{90}{T.light} & \rotatebox{90}{Sign} & \rotatebox{90}{Veget.} & \rotatebox{90}{Terrain} & \rotatebox{90}{Sky} & \rotatebox{90}{Person} & \rotatebox{90}{Rider} & \rotatebox{90}{Car} & \rotatebox{90}{Truck} & \rotatebox{90}{Bus} & \rotatebox{90}{Train} & \rotatebox{90}{M.bike} & \rotatebox{90}{Bike} & mIoU$^{16}$ \\
\midrule
ADVENT
& 85.6 & 42.2    & 79.7     & 8.7  & 0.4   & 25.9 & 5.4   & 8.1  & 80.4 & - & 84.1 & 57.9   & 23.8  & 73.3 & - & 36.4 & - & 14.2 & 33.0 & 41.2 \\
CBST
& 68.0 & 29.9    & 76.3     & 10.8 & 1.4   & 33.9 & 22.8  & 29.5 & 77.6 & - & 78.3 & 60.6   & 28.3  & 81.6 & - & 23.5 & - & 18.8 & 39.8 & 42.6 \\
CAG\_DA 
& 84.7 & 40.8    & 81.7     & 7.8  & 0.0   & 35.1 & 13.3  & 22.7 & 84.5 & - & 77.6 & 64.2   & 27.8  & 80.9 & - & 19.7 & - & 22.7 & 48.3 & 44.5 \\
PIT  
& 83.1 & 27.6    & 81.5     & 8.9  & 0.3   & 21.8 & 26.4  & {33.8}	& 76.4 & - & 78.8 & 64.2   & 27.6  & 79.6 & - & 31.2 & - & 31.0 & 31.3 & 44.0 \\
FADA 
& 84.5 & 40.1    & 83.1     & 4.8  & 0.0   & 34.3 & 20.1  & 27.2 & 84.8 & - & 84.0 & 53.5   & 22.6  & 85.4 & - & 43.7 & - & 26.8 & 27.8 & 45.2 \\
PyCDA
& 75.5 & 30.9    & 83.3     & 20.8 & 0.7   & 32.7 & 27.3  & 33.5 & 84.7 & - & 85.0 & 64.1   & 25.4  & 85.0 & - & 45.2 & - & 21.2 & 32.0  & 46.7 \\
IAST  
& 81.9 & 41.5    & 83.3     & 17.7 & 4.6   & 32.3 & 30.9  & 28.8 & 83.4 & - & 85.0 & 65.5   & 30.8  & 86.5 & - & 38.2 & - & 33.1 & 52.7 & 49.8 \\
SAC   
& \textbf{89.3} & \textbf{47.2}   & 85.5    & 26.5 & 1.3   & \underline{43.0} & 45.5  & 32.0 & \textbf{87.1} & - & 89.3	& 63.6   & 25.4  & 86.9 & - & 35.6 & - & 30.4 & 53.0 & 52.6 \\
ProDA   
& 87.1 & 44.0   & 83.2    & 26.9 & 0.7   & 42.0 & 45.8  & 34.2 & 86.7 & - & 81.3	& 68.4   & 22.1  & 87.7 & - & 50.0 & - & 31.4 & 38.6 & 51.9 \\
%
DAFormer
& 84.5 & 40.7   & \textbf{88.4}    & \underline{41.5} & \textbf{6.5}   & \textbf{50.0} & \underline{55.0}  & \textbf{54.6} & 86.0 & - & \underline{89.8}    & \underline{73.2}   & \underline{48.2}  & 87.2 & - & \underline{53.2} & - & \underline{53.9} & \underline{61.7} & \underline{60.9}  \\ 
\midrule
\method 
& 85.3 & \underline{45.7}   & \underline{87.6}    & \textbf{42.8} & \underline{5.0}   & 41.9 & \textbf{57.8}  & \underline{49.4} & \underline{86.8} & - & \textbf{89.9}    & \textbf{75.8}   & \textbf{49.0}  & \underline{88.0} & - & \textbf{61.6} & - & \textbf{54.1} & \textbf{64.8} & \textbf{61.6}  \\ 
\bottomrule
\end{tabular}
\vspace{-10pt}
\end{table*}

\subsection{Experiments on Domain Adaptive Segmentation}
\label{sec:exp_da}

In this section, we evaluate the efficacy of our \method under the domain adaptation (DA) setting.
Different from semi-supervised learning, DA often suffers from a domain shift between labeled source domain and unlabeled target domain \citep{hoyer2022daformer, zhang2021prototypical, li2022class, chen2019progressive, hoffman2018cycada, li2019bidirectional, luo2019taking},
which is more important to make sufficient use of \textit{all} pseudo-labels.
We first describe experimental settings for domain adaptive semantic segmentation.
Then, we compare our \method with state-of-the-art alternatives.
Finally, we provide qualitative results.

\paragraph{Datasets.}
In this setting, we use synthetic images from either GTA5 \citep{richter2016gta} or SYNTHIA \citep{ros2016synthia} as the source domain and use real-world images from Cityscapes \citep{cityscapes} as the target domain.
GTA5 \citep{richter2016gta} consists of $24,996$ images with resolution $1914\times 1052$, and SYNTHIA \citep{ros2016synthia} contains $9,400$ images with resolution $1280\times 760$.

\paragraph{Network Structure.}
For \textit{DA segmentation}, we use MiT-B5 \citep{xie2021segformer} pre-trained on ImageNet-1K \citep{imagenet} as the backbone and DAFormer \citep{hoyer2022daformer} as the decoder.
The representation head is the same that in the semi-supervised setting.

\paragraph{Evaluation Metric.}
For the DA segmentation task, we report results on the Cityscapes \texttt{val} set.
Particularly, on SYNTHIA $\to$ Cityscapes DA benchmark, 16 of the 19 classes of Cityscapes are used to calculate mIoU, following the common practice \citep{hoyer2022daformer}.

\begin{table}[t]
\centering
\caption{
\textbf{Ablation study on using pseudo pixels with different reliability as \textit{negative keys}}.
The reliability is measured by the entropy of pixel-wise prediction (see Sec.~\ref{sec:contra}). 
``U'' denotes \textit{unreliable}, which takes pixels with the \textit{top 20\%} highest entropy scores as negative candidates.
``R'' indicates \textit{reliable}, and means the \textit{bottom 20\%} counterpart. 
$\mathcal{D}_l$ and $\mathcal{D}_u$ are the labeled source domain and the unlabeled target domain, respectively.
The best performances are highlighted in \textbf{bold} font and the second scores are \underline{underlined}.
}
\label{tab:abalation_reliable_da}
\begin{tabular}{ccccc}
\toprule
\multirow{2}{*}{$\mathcal{D}_l$} & \multicolumn{2}{c}{$\mathcal{D}_u$} & GTA5 $\to$ Cityscapes & SYNTHIA $\to$ Cityscapes \\
\cmidrule(lr){2-3} 
\cmidrule(lr){4-4}
\cmidrule(lr){5-5}
& R & U & mIoU & mIoU$^{16}$ \\
\midrule
\checkmark & & & 68.4 & 60.8 \\
 & \checkmark & & 68.6 & 60.9 \\
 & & \checkmark & \underline{69.3} & \underline{61.2} \\
\checkmark & \checkmark & & 68.7 & 60.9 \\
\checkmark & & \checkmark & \textbf{69.6} & \textbf{61.6} \\
\checkmark & \checkmark & \checkmark & 68.5 & 60.8 \\
\bottomrule
\end{tabular}
\vspace{-10pt}
\end{table}

\begin{table}[t]
\centering
\caption{
\textbf{Performance on \textit{tailed} classes under domain adaptation settings.}
The results are averaged over 3 random seeds.
\textit{\method contributes more to tailed classes.}
}
\setlength{\tabcolsep}{2.2pt}
\label{tab:tail}
\begin{tabular}{l ccccccccc c}
\toprule
\multicolumn{11}{c}{GTA5 $\to$ Cityscapes} \\
\midrule
Method &
\rotatebox{90}{Wall} & \rotatebox{90}{T.light} & \rotatebox{90}{Sign} & \rotatebox{90}{Rider} & \rotatebox{90}{Truck} & \rotatebox{90}{Bus} & \rotatebox{90}{Train} & \rotatebox{90}{M.bike} & \rotatebox{90}{Bike} & mIoU \\
\midrule
DAFormer
& \textbf{53.5} & 55.8 & 59.4   & \textbf{44.7} & 74.5  & \textbf{78.2} & 65.1   & 55.9      & 61.8 & 60.9 \\ 
\method 
& 53.3 & \textbf{59.1}  & 58.3 & 44.6 & \textbf{77.5}  & 77.8 & \textbf{76.7}   & \textbf{59.3}      & \textbf{67.5} & \textbf{63.8} \\ 
%
%
%
\bottomrule
\end{tabular}
\vspace{-10pt}
\end{table}

\begin{figure*}[t]
    \centering
    \includegraphics[width=0.9\textwidth]{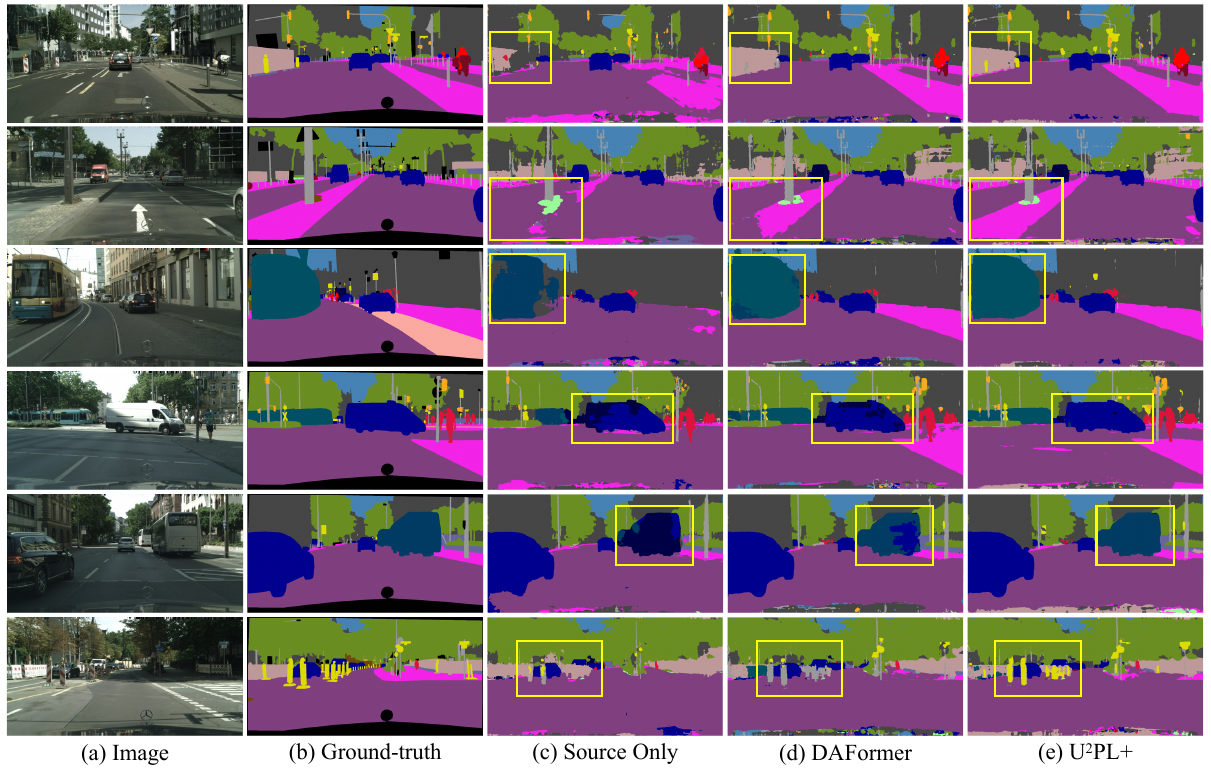}
    \vspace{-5pt}
    \caption{
    Qualitative results on \textbf{Cityscapes} \texttt{val} set under \textbf{GTA5 $\to$ Cityscapes} DA benchmark.
    (a) Input images. 
    (b) Hand-annotated labels for the corresponding image. 
    (c) \textit{Only} labeled source images are used for training.
    (d) Predictions from DAFormer \citep{hoyer2022daformer}.
    (e) Predictions from \method.
    Yellow rectangles highlight the promotion by adequately using unreliable pseudo-labels.
    }
    \label{fig:visual_gta2cs}
    \vspace{-10pt}
\end{figure*}

\paragraph{Implementation Details.}
Following previous methods \citep{zhang2021prototypical, hoyer2022daformer}, we first resize the images to $1024\times512$ for Cityscapes and $1280\times720$ for GTA5, and then randomly crop them into $512\times512$ for training.
For the testing stage, we just resize the images to $1024\times512$.
We use AdamW \citep{loshchilov2017adamw} optimizer with weight decay of $0.01$ and initial learning rate $lr_{\mathrm{base}}=6\times10^{-5}$ for the encoder and $lr_{\mathrm{base}}=6\times10^{-4}$ for the decoder follow \citep{hoyer2022daformer}.
In accordance with DAFormer \citep{hoyer2022daformer}, we warm up the learning rate in a linear schedule with $1.5$k steps and linear decay since then.
In total, the model is trained with a batch of one $512\times512$ labeled source image and one $512\times512$ unlabeled target image for $40$k iterations.
We adopt DAFormer \citep{hoyer2022daformer} as our baseline and apply extra contrastive learning with our sample strategy.
We evaluate our proposed \method under two DA benchmarks (\textit{i.e.}, GTA5 $\to$ Cityscapes and SYNTHIA $\to$ Cityscapes).
All DA experiments are conducted with a single Tesla A100 GPU.

\paragraph{Comparisons with State-of-the-Art Alternatives.}
We compare our proposed \method with state-of-the-art alternatives, including adversarial training methods (AdaptSeg \citep{tsai2018learning}, CyCADA \citep{hoffman2018cycada}, FADA \citep{wang2020classes}, and ADVENT \citep{vu2019advent}), and self-training  methods (CBST \citep{zou2018unsupervised}, IAST \citep{mei2020instance}, CAG\_DA \citep{zhang2019category}, ProDA \citep{zhang2021prototypical}, CPSL \citep{li2022class}, SAC \citep{araslanov2021self}, and DAFormer \citep{hoyer2022daformer}).

\paragraph{GTA5 $\to$ Cityscapes.}
As shown in Tab.~\ref{tab:DA}, our proposed \method achieves the best IoU performance on 13 out of 19 categories, and the IoU scores of the remaining 6 classes rank second.
Its mIoU is $69.6\%$ and outperforms DAFormer \citep{hoyer2022daformer} by $1.3\%$.
Since DAFormer \citep{hoyer2022daformer} is our baseline, the results prove that mining all unreliable pseudo-labels do help the model.
Note that there is no need to apply knowledge distillation as ProDA \citep{zhang2021prototypical} and CPSL \citep{li2022class} do, which incarnates the efficiency and simplicity of our method and the robustness of our core insight.
From Tab.~\ref{tab:DA}, we can tell that by introducing unreliable pseudo-labels into training, \method attains $76.7\%$ IoU on class \texttt{train}, where ProDA \citep{zhang2021prototypical}, FADA \citep{wang2020classes}, and ADVENT \citep{vu2019advent} even fail on this difficult long-tailed class.

\paragraph{SYNTHIA $\to$ Cityscapes.}
This DA task is more challenging than GTA5 $\to$ Cityscapes due to the larger domain gap between the labeled source domain and the unlabeled target domain.
As shown in Tab.~\ref{tab:DA}, our proposed \method achieves the best IoU performance on 8 out of 16 categories, and the mIoU over 16 classes (mIoU$^{16}$) of our method outperforms other state-of-the-art alternatives, achieving $61.6\%$.
Specifically, \method performs especially great on class \texttt{bus}, which surpasses DAFormer \citep{hoyer2022daformer} by a large margin of $8.4\%$.

\paragraph{Effectiveness of Using Unreliable Pseudo-Labels.}
We study our core insight, \textit{i.e.}, using unreliable pseudo-labels to promote domain adaptive semantic segmentation, we conduct experiments about selecting negative candidates with different reliability in Tab.~\ref{tab:abalation_reliable_da}.
As illustrated in the table, containing features from the source domain $\mathcal{D}_l$ brings marginal improvements.
Significant improvements are brought because of the introduction of unreliable predictions.

{\color{black}
\paragraph{Performance on Tailed Classes.}
From Tab.~\ref{tab:tail}, we can tell that the improvements of our \method on tailed classes are much more significant than those of all categories.
For instance, on the GTA5 $\to$ Cityscapes benchmark, \method outperforms DAFormer~\citep{hoyer2022daformer} by +1.3 mIoU (69.6 \textit{v.s.} 68.3) and \textbf{+2.9} mIoU$_{tail}$ (63.8 \textit{v.s.} 60.9), respectively.
}

\paragraph{Qualitative Results.}
Fig.~\ref{fig:visual_gta2cs} shows the improvements in segmentation results using different methods.
Benefiting from using unreliable pseudo-labels, the model is able to correct the wrong predictions when using only reliable pseudo-labels (\textit{e.g.}, DAFormer \citep{hoyer2022daformer}).
%
%

\paragraph{Discussion.}
Both DA and SS aim at leveraging a large amount of unlabeled data, while DA often suffers from domain shift, and is more important to learn domain-invariant feature representations, building a cross-category-discriminative feature embedding space.
Although our \method mainly focuses on semi-supervised learning and does not pay much attention to bridging the gap between source and target domains, using unreliable pseudo-labels provides another efficient way for generalization.
This might be because, in DA, the quality of pseudo-labels is much worse than those in SS due to the limited generalization ability of the segmentation model, making sufficient use of all pixels remains a valuable issue hence.

\subsection{Experiments on Weakly Supervised Segmentation}
\label{sec:exp_weakly}

\begin{table}[t]
    \centering
    \caption{
    Comparison with state-of-the-art methods on weakly supervised semantic segmentation benchmark.
    We report the mIoU on PASCAL VOC 2012 \texttt{val} set and \texttt{test} set.
    ``Sup.'' indicates the supervision type.
    $\mathcal{F}$, $\mathcal{I}$, and $\mathcal{S}$ represent full supervision, image-level supervision, and saliency supervision, respectively.
    $\dag$ indicates our implementation.
    }
    \label{tab:sota_ws}
    \setlength{\tabcolsep}{6pt}
    \begin{tabular}{lclccc}
    \toprule
    Method & Sup. & Backbone & \texttt{val} & \texttt{test} \\
    \midrule
    \multicolumn{5}{l}{\textit{Fully Supervised Methods}} \\
    DeepLab & \multirow{3}{*}{$\mathcal{F}$} & ResNet-101 & 77.6 & 79.7 \\
    WideResNet38 & & WR-38 & 80.8 & 82.5 \\
    SegFormer$^{\dag}$ & & MiT-B1 & 78.7 & $-$ \\
    \midrule
    \multicolumn{5}{l}{\textit{Multi-Stage Weakly Supervised Methods}} \\
    OAA+ & \multirow{5}{*}{$\mathcal{I} + \mathcal{S}$} & ResNet-101 & 65.2 & 66.4 \\
    MCIS & & ResNet-101 & 66.2 & 66.9 \\ 
    AuxSegNet & & WR-38 & 69.0 & 68.6 \\
    NSROM & & ResNet-101 & 70.4 & 70.2 \\
    EPS & & ResNet-101 & \textbf{70.9} & \textbf{70.8} \\
    \midrule
    SEAM & \multirow{6}{*}{$\mathcal{I}$} & WR-38 & 64.5 & 65.7 \\
    SC-CAM & & ResNet-101 & 66.1 & 65.9 \\
    CDA & & WR-38 & 66.1 & 66.8 \\
    AdvCAM & & ResNet-101 & 68.1 & 68.0 \\ 
    CPN & & ResNet-101 & 67.8 & 68.5 \\ 
    RIB & & ResNet-101 & \textbf{68.3} & \textbf{68.6} \\
    \midrule
    \multicolumn{5}{l}{\textit{Single-Stage Weakly Supervised Methods}} \\
    EM & \multirow{7}{*}{$\mathcal{I}$} & VGG-16 & 38.2 & 39.6 \\
    MIL & & $-$ & 42.0 & 40.6 \\
    CRF-RNN & & VGG-16 & 52.8 & 53.7 \\
    RRM & & WR-38 & 62.6 & 62.9 \\
    1Stage & & WR-38 & 62.7 & 64.3 \\
    AFA$^{\dag}$ & & MiT-B1 & 64.9 & 66.1 \\
    \method & & MiT-B1 & \textbf{66.4} & \textbf{67.0} \\
    \bottomrule
    \end{tabular}
\end{table}

\begin{table}[t]
\centering
\caption{
\textbf{Ablation study on using pseudo pixels with different reliability as \textit{negative keys}}.
The reliability is measured by the scores of CAMs (see Sec.~\ref{sec:method_weakly} for details). 
Specifically, we study the formulation of $n_{ij}(c)$ defined in Eq.~(\ref{eq:negative_ws}), which indicates whether pixel $j$ from the $i$-th image is a qualified negative key for class $c$.
Incorporating $\mathbbm{1}[\mathbf{M}_{ij}^c < \beta]$ indicates unreliable predictions for class $c$ is considered as negative keys.
The mIoU on PASCAL VOC 2012 \texttt{val} set is reported.
}
\label{tab:abalation_reliable_ws}
\setlength{\tabcolsep}{20pt}
\begin{tabular}{ccc}
\toprule
\multicolumn{2}{c}{$n_{ij}(c)$} & \multirow{2}{*}{mIoU} \\
\cmidrule(lr){1-2} 
$\mathbbm{1}[y_i^c = 0]$ & $\mathbbm{1}[\mathbf{M}_{ij}^c < \beta]$ \\
\midrule
\checkmark & & 65.0 \\
& \checkmark & 65.7 \\
\checkmark & \checkmark & \textbf{66.4} \\
\bottomrule
\end{tabular}
\vspace{-10pt}
\end{table}

In this section, we evaluate the efficacy of our \method under the weakly supervised (WS) setting.
Differently, under this setting, pixel-level annotations are entirely \textit{inaccessible}, what we have is only image-level labels,
making it crucial to make sufficient use of \textit{all} pixel-level predictions.
We first describe experimental settings for domain adaptive semantic segmentation.
Then, we compare our \method with state-of-the-art alternatives.
Finally, we provide qualitative results.

Several WS methods with image-level labels adopt a multi-stage framework, \textit{e.g.}, \citep{jiang2019integral, sun2020mining}.
They first train a classification model and  generate CAMs to be pseudo-labels.
Additional refinement techniques are usually required to improve the quality of these pseudo-labels.
Finally, a standalone semantic segmentation network is trained using these pseudo-labels.
This type of training pipeline is obviously not efficient.
To this end, we select a representative \textit{single-stage} method, AFA \citep{ru2022learning} as the baseline, where the final segmentation model is trained end-to-end.

\paragraph{Datasets.}
In this setting, we conduct experiments on PASCAL VOC 2012 \citep{voc}, which contains 21 semantic classes (including the background
class).
Following common practices \citep{fan2020cian, fan2020employing, fan2020learning}, it is  augmented with the SBD dataset \citep{sbd}, resulting in $10,582$, $1,449$, and $1,464$ images for training, validation, and testing, respectively. 

\paragraph{Network Structure.}
We take AFA \citep{ru2022learning} as the baseline, which uses the Mix Transformer (MiT) \citep{xie2021segformer} as the backbone.
A simple MLP head is used to be the segmentation head following \citep{xie2021segformer}.
The backbone is pre-trained on ImageNet-1K \citep{imagenet}, while other parameters are randomly initialized.
The representation head is the same as in the semi-supervised setting.

\begin{figure}[t]
    \centering
    \includegraphics[width=1\linewidth]{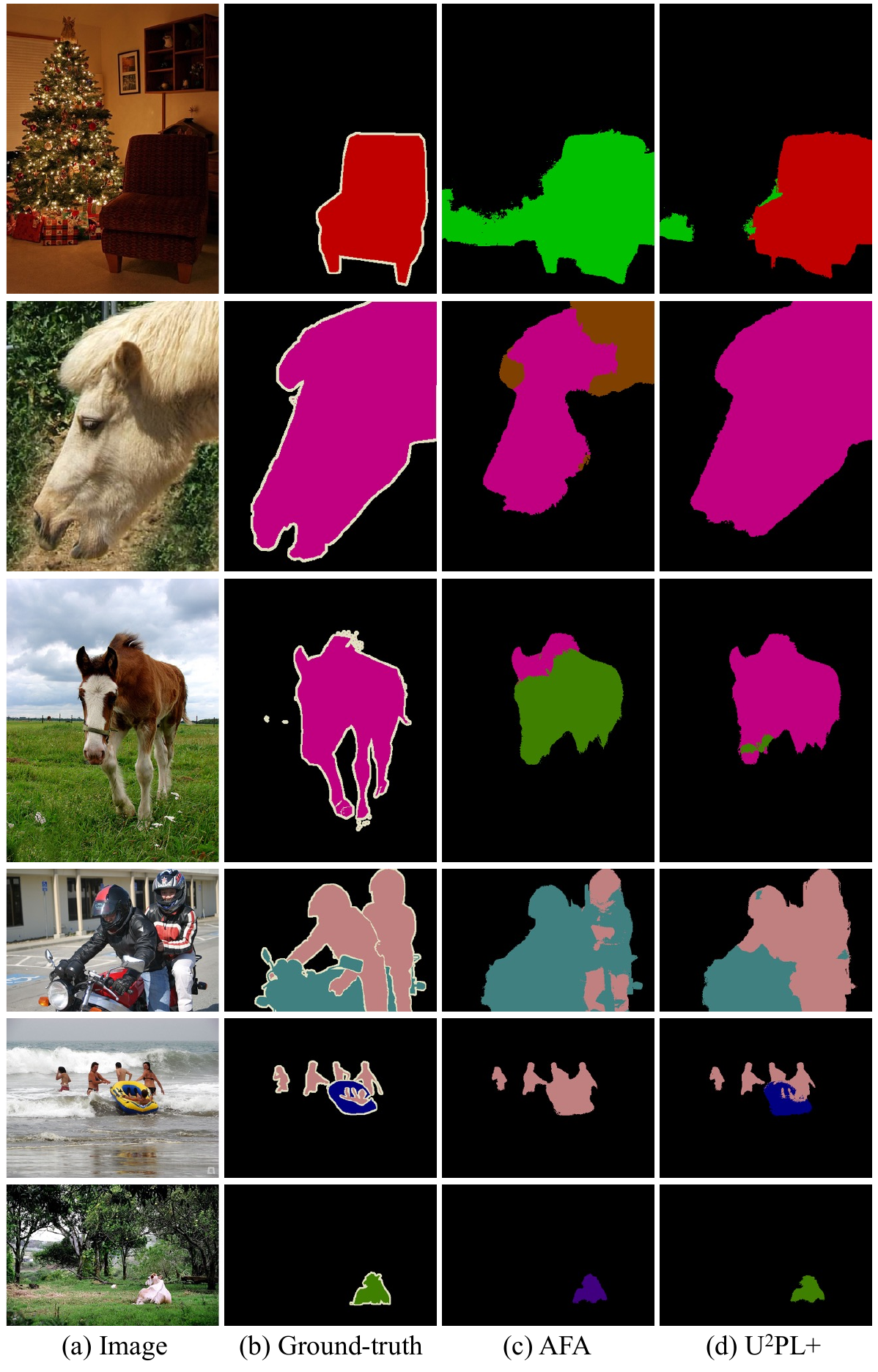}
    \vspace{-15pt}
    \caption{
    Qualitative results on PASCAL VOC 2012 \texttt{val} set under the weakly supervised setting.
    (a) Input images. 
    (b) Labels for the corresponding image. 
    (c) Predictions from our baseline AFA \citep{ru2022learning}.
    (d) Segmentation maps predicted by our \method.
    }
    \label{fig:visual_ws}
    \vspace{-10pt}
\end{figure}

\paragraph{Evaluation Metric.}
By default, we report the mIoU on \textit{both} the validation set and testing set as the evaluation criteria.

\paragraph{Implementation Details.}
Following the standard configuration of AFA \citep{ru2022learning}, the AdamW optimizer is used to train the model, whose initial learning rate is set to $6\times 10^{-5}$ and decays every iteration with a polynomial scheduler for parameters of the backbone.
The learning rates for other parameters are ten times that of the backbone.
The weight decay is fixed at $0.01$.
Random rescaling with a range of $[0.5, 2.0]$, random horizontal flipping, and random cropping with a cropping size of $512\times 512$ are adopted.
The network for $20,000$ iterations with a batch size of $8$.
All WS experiments are conducted with $2$ 2080Ti GPUs.

\paragraph{Comparisons with State-of-the-Art Alternatives.}
In Tab.~\ref{tab:sota_ws}, we compare our proposed \method with a wide range of representative state-of-the-art alternatives, including multi-stage methods and single-stage methods.
Some multi-stage methods further leverage saliency maps, including OAA+ \citep{jiang2019integral}, MCIS \citep{sun2020mining}, AuxSegNet \citep{xu2021leveraging}, NSROM \citep{yao2021non}, and EPS \citep{lee2021railroad}, while others do not, including SEAM \citep{wang2020self}, SC-CAM \citep{chang2020weakly}, CDA \citep{su2021context}, AdvCAM \citep{lee2021anti}, CPN \citep{zhang2021complementary}, and RIB \citep{lee2021reducing}.
Single-stage methods include EM \citep{papandreou2015weakly}, MIL \citep{pinheiro2015image}, CRF-RNN \citep{roy2017combining}, RRM \citep{zhang2020reliability}, 1Stage \citep{araslanov2020single}, and AFA \citep{ru2022learning}.
All methods are refined using CRF.

It has been illustrated in Tab.~\ref{tab:sota_ws} that \method brings improvements of $+1.5\%$ mIoU and $+0.9\%$ mIoU over AFA \citep{ru2022learning} on the \texttt{val} set and the \texttt{test} set, respectively.
\method clearly surpasses previous state-of-the-art single-stage methods with significant margins.
It is worth noticing that \method manages to achieve competitive results even when compared with multi-stage methods, \textit{e.g.}, OAA+ \citep{jiang2019integral}, MCIS \citep{sun2020mining}, SEAM \citep{wang2020self}, SC-CAM \citep{chang2020weakly} and CDA \citep{su2021context}.

\paragraph{Effectiveness of Using Unreliable Pseudo-Labels.}
We study our core insight, \textit{i.e.}, using unreliable pseudo-labels promotes weakly supervised semantic segmentation, we conduct experiments about selecting negative candidates with different reliability in Tab.~\ref{tab:abalation_reliable_ws}.
As illustrated in the table, incorporating unreliable predictions, \textit{i.e.}, $\mathbbm{1}[\mathbf{M}_{ij}^c < \beta]$, brings significant improvements.
It is worth noticing that since we do not have \textit{any} dense annotations, and thus leveraging the image-level annotations, \textit{i.e.}, $\mathbbm{1}[y_i^c = 0]$, in selecting negative candidates becomes crucial for a stable training procedure.

\paragraph{Qualitative Results.}
Fig.~\ref{fig:visual_ws} shows the improvements in segmentation results using different methods.
Those predictions are refined by CRF \citep{krahenbuhl2011efficient}.
Benefiting from using unreliable pseudo-labels, the model is able to classify ambiguous regions into correct classes.

%% file: 5.conclusion.tex
\section{Conclusion}\label{sec:conclusion}

In this paper, we extend our original U$^2$PL to \method, a \textit{unified} framework for label-efficient semantic segmentation, by including unreliable pseudo-labels into training.
Our main insight is that those unreliable predictions are usually just confused about a few classes and are confident enough \textit{not} to belong to the remaining remote classes.
\method outperforms many existing state-of-the-art methods in both semi-supervised, domain adaptive, and weakly supervised semantic segmentation, suggesting our framework provides a new promising paradigm in label-efficient learning research.
Our ablation experiments prove the insight of this work is quite solid, \textit{i.e.}, only when introducing unreliable predictions into the contrastive learning paradigm brings significant improvements under both settings.
Qualitative results give visual proof of its effectiveness, especially the better performance on borders between semantic objects or other ambiguous regions.
